\documentclass[letterpaper,journal]{IEEEtran}
\usepackage{amsmath,amsfonts}
\usepackage{algorithmic}
\usepackage{algorithm}
\usepackage{array}
\usepackage[caption=false,font=normalsize,labelfont=sf,textfont=sf]{subfig}
\usepackage{textcomp}
\usepackage{stfloats}
\usepackage{url}
\usepackage{verbatim}
\usepackage{graphicx}
\usepackage{cite}
\usepackage[colorlinks,bookmarks=false,citecolor=blue,linkcolor=blue,urlcolor=blue]{hyperref}
\hypersetup{
  pdftitle={Adaptive Interaction Network for Human Motion Prediction During Human-Robot Collaboration},
  pdfauthor={Mengyuan Liu, Yangting Lin, Qiongjie Cui},
  pdfsubject={Author accepted manuscript; DOI: 10.1109/TIP.2026.3718431}
}
\usepackage{orcidlink}
\usepackage{bm}
\usepackage{booktabs}
\usepackage{multirow}
\usepackage{makecell}
\usepackage[table]{xcolor}
\begin{document}

\title{Adaptive Interaction Network for Human Motion Prediction During Human-Robot Collaboration}

\author{Mengyuan~Liu~\orcidlink{0000-0002-6332-8316},~Yangting~Lin,~Qiongjie~Cui~\orcidlink{0000-0002-8078-6706}%
\thanks{
This work was supported by National Natural Science Foundation of China (No. 62473007), Guangdong Provincial Key Laboratory of Ultra High Definition Immersive Media Technology (No. 2024B1212010006), Shenzhen Innovation in Science and Technology Foundation for The Excellent Youth Scholars (No. RCYX20231211090248064). \emph{Corresponding author: Yangting Lin (e-mail: yangtinglin@stu.pku.edu.cn).}
Mengyuan Liu and Yangting Lin are with the State Key Laboratory of General Artificial Intelligence, Peking University, Shenzhen Graduate School, Shenzhen, Guangdong 518055, China, and also with the Guangdong Provincial Key Laboratory of Ultra High Definition Immersive Media Technology, Peking University, Shenzhen Graduate School, Shenzhen, Guangdong 518055, China.
Qiongjie Cui is with the National Key Laboratory of Human-Machine Hybrid Augmented Intelligence, Xi’an Jiaotong University, Xi’an, Shaanxi 710049, China.

This is the author's accepted manuscript. The Version of Record was published in \emph{IEEE Transactions on Image Processing}, vol.~35, pp.~8484--8496, 2026, and is available at \url{https://doi.org/10.1109/TIP.2026.3718431}.

\textcopyright~2026 IEEE. Personal use of this material is permitted. Permission from IEEE must be obtained for all other uses, in any current or future media, including reprinting/republishing this material for advertising or promotional purposes, creating new collective works, for resale or redistribution to servers or lists, or reuse of any copyrighted component of this work in other works.
}}%

\markboth{IEEE Transactions on Image Processing}
{Liu \MakeLowercase{\textit{et al.}}: Adaptive Interaction Network for Human Motion Prediction During Human-Robot Collaboration}

\maketitle
\begin{abstract}
Human motion prediction during human-robot collaboration is critical for achieving safe and efficient interactions in shared environments. Unlike previous studies that have  primarily focused on humans and objects, we focus on the robot-aware human motion prediction task, which explicitly models the influence of robots on human motion. This task presents new challenges, such as capturing human-robot heterogeneity and modeling complex spatial interactions. To address these issues, we develop an adaptive interaction network (AINet) model that consists of two branches: a primary branch that predicts future human motion and an auxiliary branch that estimates robot trajectories. The two branches are jointly optimized and coupled via a Local-Global Spatial Interaction (LGSI) Module, which effectively captures fine-grained and global contextual dependencies between human and robot motion sequences. Then, we introduce an Adaptive Weighted Aggregation (AWA) Module to dynamically fuse motion features using context-dependent weights, thereby increasing adaptability across diverse scenarios. Furthermore, we adopt a coarse-to-fine prediction strategy, in which a coarse pseudo-trajectory is first predicted, followed by the refinement of detailed human poses based on that trajectory. Extensive experiments on three challenging datasets reveal that our proposed method achieves superior performance, validating its effectiveness in human-robot collaboration scenarios. 
Our code is available at: \url{https://github.com/LyTingHub/AINet}.
\end{abstract}


\section{Introduction}
\IEEEpubidadjcol

\IEEEPARstart{H}{uman} motion prediction (HMP) aims to predict future human movements based on historical motion sequences. HMP has garnered increasing attention due to its wide range of applications across numerous domains, such as autonomous vehicles~\cite{10721198, wang2023learning, wang2023dynamic}, pedestrian tracking~\cite{zhang2024decouple, chen2022multimodal, 9268972} and robotics~\cite{unhelkar2018human, salzmann2023robots}. Specifically, HMP is particularly crucial as it enables robots to anticipate human movements and responsively adapt their motion planning, thereby ensuring safe and efficient collaboration in shared workspaces~\cite{li2021multiscale, lee2016parallel, cai2022human, meng2024hierarchical, 10570381, 9763877}.

\begin{figure}[!t]
\centering
\includegraphics[width=3.2in]{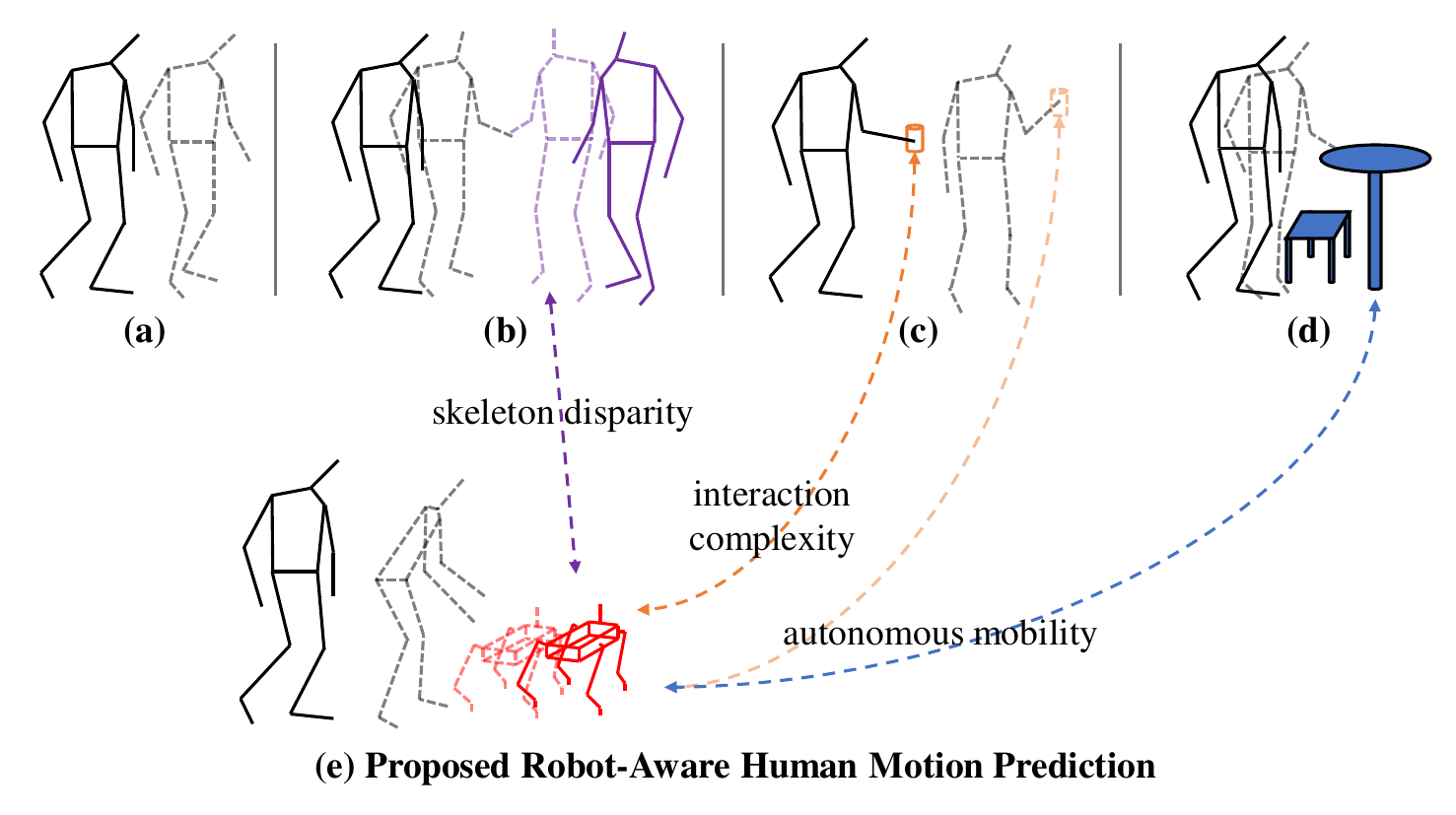} 
\vspace{-1.4em}
\caption{Comparison of HMP paradigms. The solid and dashed human skeletons represent past observations and future motions, respectively: (a) single-person motion prediction\cite{mao2019learning}, (b) multi-person motion prediction\cite{guo2022multi}, (c) HMP during human-object interactions\cite{corona2020context}, (d) scene-aware HMP\cite{cao2020long}, and (e) our proposed robot-aware HMP. The proposed setting introduces unique challenges: (1) inherent skeleton disparity between humans and robots, (2) interaction complexity due to the robot's structure and mobility, and (3) dynamic context variations caused by the robot's autonomous mobility. }
\vspace{-2.76em}
\label{motivation}
\end{figure}

{
However, most existing HMP methods are not designed for robot-aware settings. Prior approaches can be broadly categorized into single-person~\cite{mao2019learning}, multi-person~\cite{guo2022multi}, human–object interaction (HOI)~\cite{corona2020context}, and scene-aware~\cite{cao2020long} methods. While effective in their respective domains, they fail to generalize to HRC due to three fundamental mismatches, as illustrated in Fig.~\ref{motivation}. First, single-person methods neglect inter-agent coupling and cannot model mutual influence. Second, multi-person approaches assume homogeneous agents with symmetric interaction patterns, which is incompatible with the heterogeneous embodiments and roles of humans and robots. Third, HOI and scene-aware methods treat context as passive or static, whereas a robot is an autonomous and reactive agent whose motion dynamically affects human behavior.
}

{
These limitations highlight three key challenges for robot-aware HMP: (1) \textit{kinematic heterogeneity} between human and robot embodiments, (2) \textit{bidirectional and asymmetric coupling} between agents, and (3) \textit{dynamically evolving context} induced by robot autonomy. Addressing these challenges requires a fundamentally different modeling paradigm that explicitly accounts for both agents and their interactions.
}

    To tackle these challenges, we introduce 
Adaptive Interaction Network (AINet) that explicitly models these asymmetric and dynamic interactions.
Given that robots possess autonomous mobility, which actively alters environmental contexts and introduces dynamic uncertainties in human-robot cooperation, we adopt a dual-branch architecture that models human and robot motions separately. This framework not only predicts human motion as the primary task but also incorporates robot motion prediction as an auxiliary task, enabling comprehensive modeling of both human and robot behaviors to address the complexities of shared environments.

{
To comprehensively model human-robot interaction, we design our method around the fact that coordination in shared spaces relies on spatial dependencies at different semantic scales. A single interaction representation is insufficient: local cues support contact-sensitive adaptations, while global states capture overall coordination trends. Thus, we propose an attention-based Local-Global Spatial Interaction (LGSI) Module to model and integrate these dual-scale cues.
To address structural heterogeneity between human and robot embodiments, LGSI adopts an Align-Interact-Reconstruct pipeline. It projects features into a shared latent space for cross-attention modeling, then reconstructs them back to their respective domains, preserving embodiment-specific semantics while capturing cross-agent dependencies.
}

{
Since the importance of local versus global cues varies by scenario (e.g., near-contact vs. loose coordination), using fixed fusion coefficients is suboptimal. Therefore, we introduce an Adaptive Weighted Aggregator (AWA) to dynamically fuse local and global features using context-dependent weights.
}

{
Finally, we adopt a coarse-to-fine prediction strategy because global tendencies and local poses play fundamentally different roles in motion generation. To resolve long-horizon inconsistencies without relying on an oversimplified, fixed hip-joint proxy, we introduce the Traj-Pose Module. It learns a pseudo-trajectory that adaptively aggregates multi-joint motion cues. Specifically, an ST-Encoder captures spatio-temporal dynamics, a Traj Module generates the global pseudo-trajectory, and a Pose Module refines future poses conditioned on it.
}

We conduct evaluations on two HRC datasets, including the CoMaD~\cite{kedia2024interact} dataset, which focuses on close-proximity HRC in domestic kitchens, and the CHICO~\cite{sampieri2022pose} dataset, which is related to HRC in factory assembly lines. In addition, we test the effectiveness of our method in human-robot interaction (HRI) scenarios using the HARPER~\cite{avogaro2024exploring} dataset, which involves interactions between a human and a quadruped robot at various distances. Our method has demonstrated superior performance on these datasets, highlighting its potential for real-world applications in human-robot collaboration.

Our contributions can be summarized as follows:
\begin{itemize}
\item {To effectively model complex interaction patterns, we propose a Local-Global Spatial Interaction (LGSI) that decomposes interaction information into fine-grained local details and coarse-grained global representations, enabling robust performance in complex scenarios.} 
\item {For enhanced interaction dynamics modeling, we develop an Adaptive Weighted Aggregator (AWA) that intelligently fuses multi-grained motion features using context-based weights, improving the model's ability to capture subtle interaction patterns.}
\item {We introduce a Traj-Pose Module that separates global motion tendency from local pose refinement via a coarse-to-fine strategy and a learned pseudo-trajectory, avoiding limitations of fixed hip-based trajectory representations.}
\item {Our approach achieves state-of-the-art performance across three challenging HMP datasets for human-robot collaboration and interaction: CoMaD~\cite{kedia2024interact}, CHICO\cite{sampieri2022pose} and HARPER~\cite{avogaro2024exploring} datasets.
}
\end{itemize}

\section{Related Work}
\subsection{Human Motion Prediction}

Early work primarily focused on predicting future poses of a single human given past observations~\cite{martinez2017human, mao2020history, guo2023back, wang2024gcnext,wang2024skeleton, tian2024transfusion}. These methods leveraged RNNs~\cite{martinez2017human, 9321130}, temporal convolutions~\cite{mao2020history}, or MLPs~\cite{guo2023back} to model motion dynamics efficiently.
However, they assumed that the human moves in isolation and ignore external contextual cues, limiting their applicability in real-world scenarios that involve collaboration or shared environments.

To address the limitations of single-person models in interactive settings, subsequent research shifted to multi-person motion prediction, focusing on modeling interpersonal dynamics~\cite{peng2023trajectory, zhu2024social, medina2024context, xiao2024multi}. 
For instance, \cite{peng2023trajectory} designed a body-level transformer to capture detailed social interactions in crowds, while \cite{zhu2024social} utilized cognitive hierarchies for team sports. 
A critical limitation of these works, however, is the assumption of homogeneous agents—individuals with similar skeletal structures and symmetric behavioral logic. 
{At the algorithmic level, they rely on shared-weight encoders, symmetric self-attention mechanisms (e.g., shared queries, keys, and values across all agents), or static adjacency matrices defined strictly by standard human skeletal topologies. }
This assumption is invalid in human-robot interaction, which involves heterogeneous agents with fundamentally different kinematics and decision-making processes. Consequently, these models are inadequate for robot-aware prediction tasks.
{These architectural limitations highlight the critical need for a paradigm shift in robot-aware settings. Instead of a shared-weight homogeneous network, our proposed AINet employs a dedicated \textit{dual-branch architecture} that structurally isolates the morphological encoding of humans and robots, explicitly addressing kinematic heterogeneity before engaging in cross-agent interaction reasoning.}

HOI provides an important direction for contextual human motion prediction~\cite{corona2020context, yan2024forecasting, hu2024hoimotion, lagamtzis2023exploiting }. \cite{corona2020context} introduced a semantic-graph framework that models humans and surrounding objects as graph nodes and iteratively learns their dependencies for motion forecasting. \cite{yan2024forecasting} further extended this line to whole-body prediction in grasping scenarios by fusing body-hand dynamics with object shape features and human-object distance cues. 
{
However, these methods are designed for passive objects that are manipulated or observed by humans. Their contextual representations mainly encode object category, geometry, or relative distance, which are suitable for tools or grasped items but insufficient for robots that possess articulated motion, autonomous decision-making, and adaptive behaviors. Consequently, they cannot capture the bidirectional and asymmetric interaction dynamics required in HRC.
}

Scene-aware approaches~\cite{cao2020long, xing2024scene, lou2024multimodal} instead model the broader environment as a source of spatial or geometric constraints. \cite{cao2020long} exploited scene images in a goal-path-pose framework to guide long-term destination prediction and path planning, while \cite{xing2024scene} introduced explicit human-scene mutual distances and SDF-based scene representations to enforce geometric consistency in whole-body forecasting. 
{Although effective for static environments, these methods treat context as passive scene layout or geometry. Such representations are suitable for encoding occupancy, collision, and environmental semantics, but they cannot represent a robot as a dynamic interactive partner whose motion evolves over time and actively affects human behavior. Therefore, they are also insufficient for HRC, where prediction depends on continuous mutual adaptation between humans and autonomous robots.
}

\subsection{Human Motion Prediction for Human-Robot Collaboration}

{
HRI is an umbrella term that encompasses diverse forms of human--robot coupling, including cooperation, avoidance, physical contact, and communication, whereas HRC refers more specifically to task-oriented coordination toward a shared goal. In this paper, we primarily target robot-aware human motion prediction in HRC scenarios (e.g., CoMaD and CHICO), while also evaluating on a broader HRI benchmark (HARPER) to examine whether robot-aware modeling generalizes to interaction-rich behaviors beyond explicit collaboration.
}

{
Despite the importance of motion anticipation in close-proximity teamwork, only a limited number of works explicitly study pose forecasting in HRC settings. \cite{sampieri2022pose} introduced SeS-GCN for industrial pose forecasting and released the CHICO benchmark. SeS-GCN focuses on efficient spatio-temporal graph modeling of intra-human joint dynamics (including learned sparsity), and the robot is mainly exploited as known context in downstream safety analyses (e.g., collision checking) rather than being modeled as a dynamic agent that continuously shapes human adaptation. Consequently, the predictor is not designed to capture bidirectional and asymmetric coupling between human and robot motions. In parallel, \cite{yasar2024posetron} proposed PoseTron together with the multimodal INTERACT dataset for close-proximity collaboration. PoseTron formulates HRC as a multi-agent forecasting problem with a transformer encoder--decoder and attention mechanisms to fuse modalities, but it provides limited inductive bias for heterogeneous embodiments and role-asymmetric dependencies that are central to robot-aware HRC.
}

{
Related robotics systems often integrate human forecasts into planning objectives (e.g., cost-aware forecasting for collaborative manipulation)~\cite{11163873, 10570381, kedia2023manicast}, but they typically do not learn an explicit, coupled human-robot motion model at the pose level. Motivated by these gaps, we propose a dedicated robot-aware forecasting framework. Our dual-branch architecture models human and robot motion with separate encoders/decoders while explicitly reasoning about their interplay via a LGSI module. LGSI combines fine-grained proximity cues (critical for contact/avoidance) with coarse global interaction states (important for long-horizon coordination). An auxiliary robot prediction task further reinforces bidirectional awareness during training, enabling robust generalization across both structured HRC and interaction-rich HRI settings.
}
{
Improved pose prediction accuracy also benefits downstream robotic tasks. 
Lower MPJPE reduces the uncertainty of predicted human motion, which can decrease conservative waiting and replanning in execution, while enabling tighter safety margins in human-aware planning.
}

\begin{figure*}[!t]
\centering
\vspace{-3em}
\includegraphics[width=\textwidth]{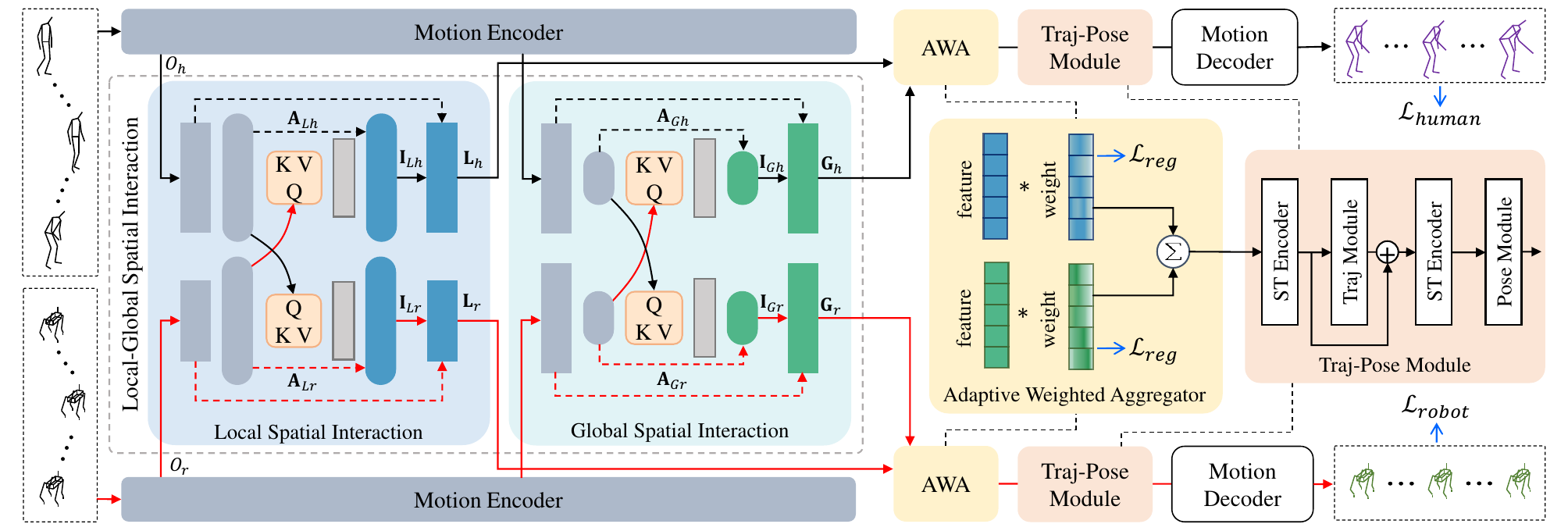}
\caption{An overview of our proposed AINet. 
The input motion data is first transformed into the frequency domain using the Motion Encoder. Human and robot features then interact through the LGSI module, which generates local and global interaction features via a three-step process: Align, Interact, and Reconstruct. The AWA module dynamically aggregates these interaction features using context-based weights. Next, the Traj-Pose Module extracts global trajectories and refines local poses. Finally, the Motion Decoder reconstructs the motion sequence from its frequency components. For training, we utilize the combined loss function \(\mathcal{L}_{human} + \mathcal{L}_{robot} + \lambda\mathcal{L}_{reg}\), where \(\mathcal{L}_{human}\) and \(\mathcal{L}_{robot}\) supervise the human and robot branches, respectively, while \(\mathcal{L}_{reg}\) regularizes the weights generated by AWA.} 
\vspace{-1.5em}
\label{overview}
\end{figure*}

\section{Adaptive Interaction Network for Robot-Aware Human Motion Prediction}
Robot-aware human motion prediction is designed to predict future human pose sequences, $\mathbf{H_F} \in \mathbb{R}^{N \times 3J}$, on the basis of previously observed human pose sequences, $\mathbf{H_P} \in \mathbb{R}^{M \times 3J}$, and observed robot pose sequences $\mathbf{R_P} \in \mathbb{R}^{M \times 3K}$. Furthermore, we predict future robot poses as an auxiliary task. In this context, $M$ and $N$ represent the number of frames in the past and future, respectively, whereas $J$ and $K$ represent the number of joints in the human and robot, respectively.

To simplify notation, we use the subscript $h$ to denote quantities related to the human and the subscript $r$ to denote quantities related to the robot. If no subscript is specified, it applies to either the human or the robot.

The overall architecture of our proposed framework is illustrated in Fig.~\ref{overview}. The network consists of two symmetric branches for predicting human and robot motion. Each branch consists of a Motion Encoder for feature extraction, a Local-Global Spatial Interaction (LGSI) module that connects both branches to capture cross-agent interactions, an Adaptive Weighted Aggregator (AWA) that dynamically fuses interaction features using context-based weights, and a Traj-Pose Module that hierarchically predicts pseudo-trajectory features followed by detailed poses. The processed features are finally decoded by a Motion Decoder to generate future motion. 

\subsection{Motion Encoder and Motion Decoder}
We leverage the Motion Encoder to perform initial feature encoding on the input motion sequences. The encoder comprises a Discrete Cosine Transform (DCT) layer that projects the input into the frequency domain for a smoother and more compact representation, as well as a stack of fully connected (FC) layers for further feature extraction. The DCT, as an invertible transformation, has been demonstrated to be highly effective for human motion prediction tasks because it characterizes temporal smoothness and reduces high-frequency noise. 
The Motion Decoder,  which serves as the final module of our network, reconstructs the output motion sequences by decoding the processed features. It utilizes FC layers followed by an inverse Discrete Cosine Transform (iDCT). Following common practices in motion prediction, our network predicts the residual between future poses and the last observed pose, enhancing training stability and prediction accuracy. 

Throughout the pipeline, we represent 3D poses in the world coordinate system rather than in local coordinate systems{, without any additional normalization or coordinate transformation}. This design facilitates explicit perception of spatial relationships between the human and the robot, which is essential for accurate interaction modeling in joint environments. 

\subsection{Local-Global Spatial Interaction}
{
Although prior motion prediction methods have achieved strong empirical performance, the lack of explicit interaction modeling often leads to physically implausible predictions in human-robot scenarios, such as penetration artifacts, missed avoidance responses, or floating motions. 
A key reason is that interaction constraints in shared spaces are inherently multi-scale: some are local and contact-sensitive, while others reflect the broader relational state between the two agents~\cite{9055436,9954217,10066256}. 
Modeling only one of these scales is insufficient. Fine-grained representations may capture precise spatial adjustments but miss the overall interaction tendency, whereas coarse-grained representations may preserve global coordination while overlooking local geometric constraints.
}

{
To overcome these limitations, we introduce the LGSI module to model human-robot spatial interactions at two complementary physical scales. In our setting, local interactions denote short-range, part-level geometric dependencies, especially around the human hand/arm, robot end-effector, and nearby shared workspace, which directly affect fine adjustments for contact, grasp alignment, and local collision avoidance. By contrast, global interactions describe the overall spatial configuration and evolution of the human-robot pair, such as approaching, separating, yielding, or entering a handover phase, which determine the high-level interaction state and motion trend. Accordingly, the LGSI module comprises Local Spatial Interaction (LSI) and Global Spatial Interaction (GSI): LSI preserves contact-sensitive local geometry, while GSI captures the holistic collaboration context.
}

\textbf{Align}: We employ MLPs to map the original skeletons into latent spaces, mitigating the inherent disparity between human and robot structures. 
Notably, LSI and GSI differ in their latent space dimensionality: LSI utilizes a higher-dimensional latent space to enrich skeleton representations beyond keypoints, while GSI employs a lower-dimensional space for abstract interaction states. The dimensionalities are determined based on the structural complexity of the input skeletons and validated through empirical performance across different interaction scenarios. 
Let $\mathbf{O}_h \in \mathbb{R}^{M \times D_h}$ and $\mathbf{O}_r \in \mathbb{R}^{M \times D_r}$ denote the human and robot motion features encoded by the Motion Encoder, respectively, where $M$ is the sequence length and $D_h$, $D_r$ are their corresponding feature dimensions. 
The alignment process is formulated as follows:
\begin{align}
\label{Align_step}
\begin{aligned}
\mathbf{A} = \text{Align}(\mathbf{O}) = \text{MLP}(\mathbf{O}), \quad \text{where } \mathbf{O} \in \{\mathbf{O}_h, \mathbf{O}_r\}
\end{aligned}
\end{align}
The aligned feature $\mathbf{A} \in \mathbb{R}^{M \times D_a}$ shares a unified dimension $D_a$ for both human and robot motions.

\textbf{Interact}: The interaction modeling is performed in the aligned latent space. We leverage cross-attention mechanisms to discover correlations between human and robot motion sequences, revealing potential interaction relationships. The processed features are further refined through FC layers. Additionally, skip connections are incorporated to preserve the own motion information of each agent. For the human branch, the interaction processes for LSI and GSI are formulated as:
\begin{align}
\begin{aligned}
   \mathbf{I}_h= \text{Interact}(\mathbf{A}_h, \mathbf{A}_r)= \text{FC}(\text{MHA}(\mathbf{A}_h, \mathbf{A}_r)) + \mathbf{A}_h ,
\end{aligned}
\end{align}
where $\mathbf{A}_h,\mathbf{A}_r\in \mathbb{R}^{M \times D_a}$ represent the aligned feature representations of human and robot, and $\mathbf{I}_h \in \mathbb{R}^{M \times D_a}$ is the output interaction feature. $\text{MHA}(\mathbf{A}_h, \mathbf{A}_r)$ denotes multi-head attention with $\mathbf{A}_r$ as query, and $\mathbf{A}_h$ as both key and value, and $\text{FC}$ indicates the fully-connected layers. 
The MHA mechanism enables the model to attend to multiple relational cues in parallel, supporting asymmetric and dynamic correlations between agents.

\textbf{Reconstruct}: To reconstruct the human and robot motion, we employ FC layers to map the processed interaction features from the latent space back to their original skeleton spaces. Additionally, skip connections are incorporated to enhance training stability. For the human branch, the reconstruction process for LSI is formulated as:
\begin{align}
    \begin{aligned}
    \mathbf{L}_h=\text{Reconstruct}(\mathbf{O}_h, \mathbf{I}_{Lh}) &= \text{FC}(\mathbf{I}_{Lh}) + \mathbf{O}_h,
    \end{aligned}
\end{align}
where $\mathbf{O}_h \in \mathbb{R}^{M \times D_h}$ represents the output of Motion Encoder, $\mathbf{I}_{Lh}\in \mathbb{R}^{M \times D_a}$ denotes the interaction features obtained from the Interact step, and $\mathbf{L}_h\in \mathbb{R}^{M \times D_h}$ is the output. The GSI reconstruction follows the same formulation.

\subsection{Adaptive Weighted Aggregator}

{
Even when local and global interaction features are both available, their relative importance is not constant across human-robot interaction scenarios. For example, when the human is close to the robot, local geometric cues may dominate because small spatial differences can determine whether the motion corresponds to contact, avoidance, or collision. In contrast, when the agents are farther apart or only loosely coordinated, the global interaction state may provide more reliable guidance than local proximity details. Therefore, simply summing the two features or using fixed fusion coefficients would impose the same interaction bias on all contexts and limit the model's adaptability.
This design intuition is consistent with prior skeleton-sequence models~\cite{8762119}, where different motions or feature components contribute unequally to downstream reasoning, motivating adaptive gating and weighting.
}

{
To accommodate these dynamically varying interaction patterns, we propose the Adaptive Weighted Aggregator (AWA), which combines the local and global interaction features from LSI and GSI using context-dependent weights. Rather than treating local and global cues as equally informative, AWA explicitly predicts how much each interaction scale should contribute under the current context.
}
\begin{equation}
    \text{AWA}(\mathbf{L}, \mathbf{G}, \mathbf{C}) = W_\alpha(\mathbf{C}) \cdot \mathbf{L} + W_\beta(\mathbf{C}) \cdot \mathbf{G} = \alpha \mathbf{L} + \beta \mathbf{G},
\end{equation}
where $\mathbf{L}, \mathbf{G} \in \mathbb{R}^{M \times D}$ denote the local and global interaction features, with $\mathbf{L} \in \{\mathbf{L}_h, \mathbf{L}_r\}$ and $\mathbf{G} \in \{\mathbf{G}_h, \mathbf{G}_r\}$ as defined in the LSI and GSI modules. $\alpha, \beta \in \mathbb{R}^{M \times 1}$ are dynamic weights computed from the interaction context $\mathbf{C}$. We construct the context vector $\mathbf{C} = [\mathbf{A}^{GSI}_h, \mathbf{A}^{GSI}_r] \in \mathbb{R}^{M \times 2D^{GSI}_a}$ by concatenating the aligned features from the Align step of GSI, as defined in Eq.~\ref{Align_step}. 
{We derive this context from the aligned features of GSI rather than LSI, since GSI alignment provides a stable, abstract representation ideal for gating global feature fusion, whereas the contact-sensitive nature of LSI features restricts their primary utility to precise local corrections.
}

First, $\mathbf{C}$ is normalized by a Layer Normalization (LN) layer for stable value ranges.
Second, the weight generation enforces $\alpha+\beta=1$ via the following equation:
\begin{align}
\begin{aligned}
    \mathbf{C}' &= \text{LN}(\mathbf{C}), \\
    \alpha' &= \min(\max(\text{MLP}_{\alpha}(\mathbf{C}'), 0), 1)\in \mathbb{R}^{M\times1}, \\
    \beta' &= \min(\max(\text{MLP}_{\beta}(\mathbf{C}'), 0), 1)\in \mathbb{R}^{M\times1}, \\
    \alpha &= \frac{\alpha'}{\alpha' + \beta'},     \beta = \frac{\beta'}{\alpha' + \beta'}.
\end{aligned}
\end{align}

To train the AWA module, we introduce a regularization term that constrains the weight coefficients to the interval [0, 1]. Given the four weight parameters ($\alpha_h$, $\beta_h$, $\alpha_r$, $\beta_r$) for the human and robot branches, the regularization loss is formulated as follows:
\begin{equation}
\label{regloss}
    \begin{split}
        \mathcal{L}_{\text{reg}} = \sum_{w \in \{\alpha_h, \beta_h, \alpha_r, \beta_r\}} \bigl( &\| \text{ReLU}(-w) \|_1 \\
        &+ \| \text{ReLU}(w-1) \|_1 \bigr).
    \end{split}
\end{equation}
This formulation encourages the weights to remain within the interval [0, 1], which helps promote model stability and improves generalization performance.

\subsection{Traj-Pose Module}

{
Predicting future human motion in a single stage requires the model to simultaneously infer long-term movement tendency and fine-grained pose articulation, which can make learning unstable and blur the distinction between global motion structure and local pose details. We therefore adopt a coarse-to-fine strategy, in which a coarse motion representation is estimated first and detailed poses are refined afterward. This design allows the model to establish temporally coherent global dynamics before focusing on local kinematic precision.}

\begin{figure}[t]
    \centering
    \includegraphics[width=3.5in]{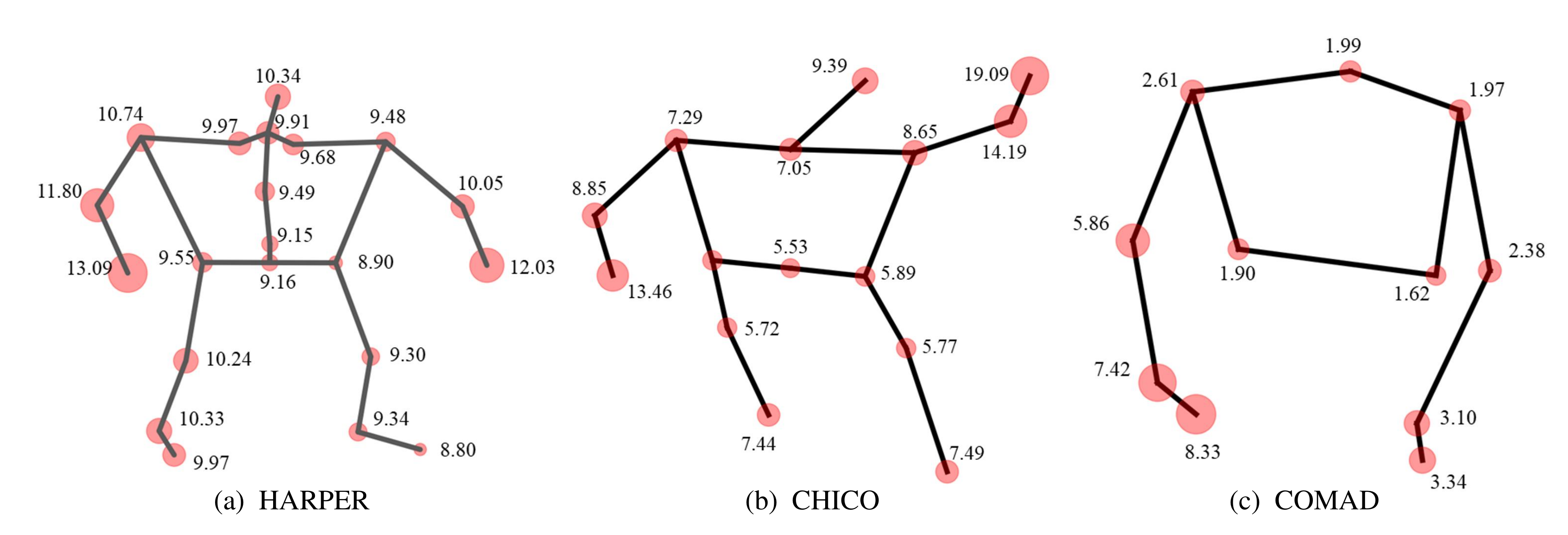}
    \vspace{-2em}
    \caption{{Joint-wise motion patterns on HARPER, CHICO, and CoMaD. Red circles indicate the average motion magnitude at each joint. (a) HARPER: hip joints exhibit large displacements; (b–c) CHICO and CoMaD: motion is primarily concentrated in the upper limbs, while hips remain nearly static. These patterns reveal the limitation of using a fixed-joint trajectory as a proxy for global motion.}} 
    \label{fig:pseudo_traj_motivation}  
    \vspace{-1em}
\end{figure}

\begin{figure}[t]
    \centering
    \includegraphics[width=3.5in]{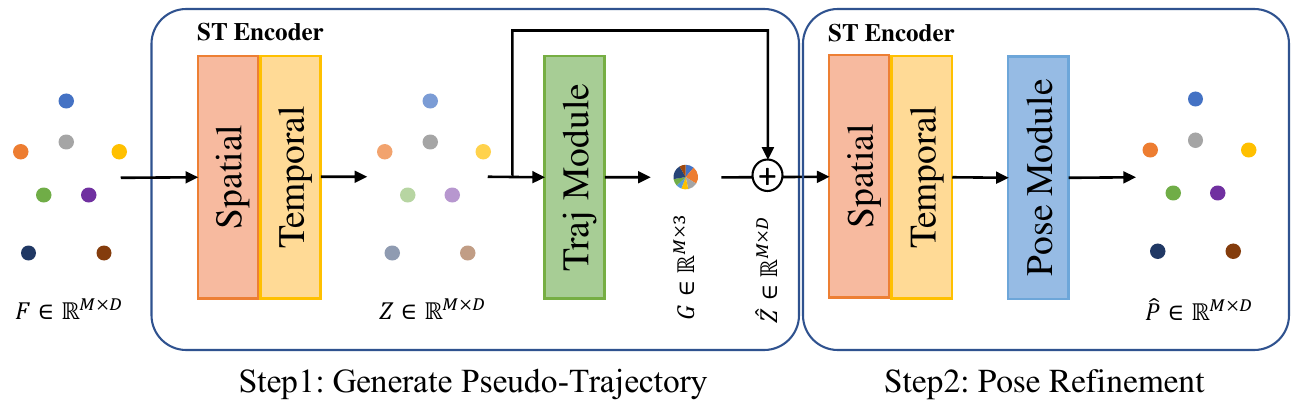}
    \vspace{-2em}
    \caption{{Architecture of the proposed Traj-Pose Module, composed of ST Encoders, a Traj Module for pseudo-trajectory generation, and a Pose Module for refinement.}} 
    \vspace{-2em}
    \label{fig:Traj_pose_module_revised}  
\end{figure}

{
A second design question is how to represent the coarse global motion. Previous methods~\cite{jeong2024multi} often use a fixed hip-joint trajectory as a proxy for global movement. However, this assumption is not always valid in human--robot interaction because the dominant motion cues can vary substantially across datasets, tasks, and embodiments. As shown in Fig.~\ref{fig:pseudo_traj_motivation}, some scenarios exhibit large root displacement, whereas others are dominated by upper-limb motion with nearly static hips. This observation suggests that a fixed-joint trajectory is too restrictive to serve as a universal global representation. 
We therefore introduce a learnable pseudo-trajectory learned from the contextualized full-body spatio-temporal representation.
}

As illustrated in Fig.~\ref{fig:Traj_pose_module_revised}, the Traj-Pose Module adopts a two-stage strategy: pseudo-trajectory generation followed by pose refinement. 
{
Given the input feature matrix $\mathbf{F} \in \mathbb{R}^{M \times D}$ from the AWA module, where $M$ denotes the number of time frames and $D=3J$ is the feature dimension for $J$ human joints, the first ST Encoder produces an intermediate representation $\mathbf{Z} = \mathrm{ST}(\mathbf{F}) \in \mathbb{R}^{M \times D}$. The Traj Module then maps the contextualized feature tensor $\mathbf{Z}$ to a low-dimensional pseudo-trajectory $\mathbf{G} = \mathcal{F}_{\text{traj}}(\mathbf{Z}) \in \mathbb{R}^{M \times 3}$, where $\mathcal{F}_{\text{traj}}(\cdot)$ is parameterized by an MLP acting along the feature dimension. To inject this global motion cue back into the feature space, we reshape $\mathbf{Z}$ as $\tilde{\mathbf{Z}} \in \mathbb{R}^{M \times J \times 3}$ and broadcast $\mathbf{G}$ along the joint dimension, yielding
\begin{equation}
    \hat{\mathbf{Z}} = \mathrm{reshape}^{-1}\!\left(\tilde{\mathbf{Z}} + \mathrm{repeat}_{J}(\mathbf{G})\right) \in \mathbb{R}^{M \times D}.
\end{equation}
The enriched representation $\hat{\mathbf{Z}}$ is then fed into the second-stage encoder for pose refinement.
}

\subsection{Training Objective}
Our training objective comprises three key components:
\begin{align}
    \mathcal{L} = \mathcal{L}_{\text{human}} + \mathcal{L}_{\text{robot}} + \lambda\mathcal{L}_{\text{reg}},
\end{align}
where $\mathcal{L}_{\text{human}}$ and $\mathcal{L}_{\text{robot}}$ denote the prediction losses for human and robot motions, respectively, and $\mathcal{L}_{\text{reg}}$ represents the regularization loss from the AWA module. The regularization term constrains the generated weights within a meaningful range, as computed in Eq.(\ref{regloss}). 

For $\mathcal{L}_{\text{human}}$ and $\mathcal{L}_{\text{robot}}$, given the predicted motion sequence $\mathbf{X} \in \mathbb{R}^{T \times Q \times 3}$ and ground truth sequence $\mathbf{Y} \in \mathbb{R}^{T \times Q \times 3}$, where $T$ denotes the temporal length and $Q$ represents the number of joints, the prediction loss is defined as follows:
\begin{equation}
\mathcal{L}_{\text{prediction}} = \|\mathbf{X} - \mathbf{Y}\|_2 + \|\mathbf{V}_{\mathbf{X}} - \mathbf{V}_{\mathbf{Y}}\|_2,
\end{equation}
where $\mathbf{V}_{\mathbf{X}} \in \mathbb{R}^{(T-1) \times Q \times 3}$ and $\mathbf{V}_{\mathbf{Y}} \in \mathbb{R}^{(T-1) \times Q \times 3}$ represent the computed velocity sequences for $\mathbf{X}$ and $\mathbf{Y}$, respectively. These velocities are obtained through temporal differentiation by applying $\mathbf{v}_t = \mathbf{x}_{t+1} - \mathbf{x}_t$ to each frame $t$, where $x_t$ represents the joint positions at time $t$.

\section{Experiments}
\label{Exps}

\begin{table}[!t]
\centering
\caption{Comparisons of the Three Datasets Used in the Experiments.}
\vspace{-1em}
\label{Dataset_compare}
\begin{tabular}{|c|ccc|}
\hline
\multirow{2}{*}{} & \multicolumn{3}{c|}{Dataset} \\ \cline{2-4} 
 & \multicolumn{1}{c|}{CoMaD} & \multicolumn{1}{c|}{CHICO} & HARPER \\ \hline
Scenario & \multicolumn{1}{c|}{HRC} & \multicolumn{1}{c|}{HRC} & HRI \\ \hline
Duration(s) & \multicolumn{1}{c|}{3,578} & \multicolumn{1}{c|}{6,773} & 2,451 \\ \hline
Robot & \multicolumn{1}{c|}{Arm} & \multicolumn{1}{c|}{Arm} & Quadruped \\ \hline
action \# & \multicolumn{1}{c|}{5} & \multicolumn{1}{c|}{7} & 15 \\ \hline
subject \# & \multicolumn{1}{c|}{9} & \multicolumn{1}{c|}{20} & 17 \\ \hline
frame \# & \multicolumn{1}{c|}{107,354} & \multicolumn{1}{c|}{169,325} & 73,530 \\ \hline
human joint \# & \multicolumn{1}{c|}{11} & \multicolumn{1}{c|}{15} & 21 \\ \hline
robot joint \# & \multicolumn{1}{c|}{10} & \multicolumn{1}{c|}{9} & 23 \\ \hline
\end{tabular}
\end{table}

In this section, we present the experimental results for three challenging datasets: CoMaD \cite{kedia2024interact}, CHICO \cite{sampieri2022pose} and HARPER~\cite{avogaro2024exploring}. 
We evaluate model performance using the Mean Per Joint Position Error (MPJPE), which is the standard metric for human motion prediction tasks. Specifically, the MPJPE measures the average Euclidean distance between the predicted joints and ground truth positions across all joints. A lower MPJPE value indicates better prediction performance.

\subsection{Datasets}
The statistics of the three benchmarks are presented in Table~\ref{Dataset_compare}. 
{Among these benchmarks, CoMaD and CHICO represent task-oriented HRC scenarios, whereas HARPER represents a broader HRI scenario. We include HARPER to evaluate whether the proposed robot-aware motion prediction framework generalizes beyond strictly collaborative settings.}
These benchmarks provide 3D pose annotations for both human and robot entities. 
We subsequently provide a brief overview of the datasets and several examples. 

\begin{figure}[!t]
\centering
\includegraphics[width=3.5in]{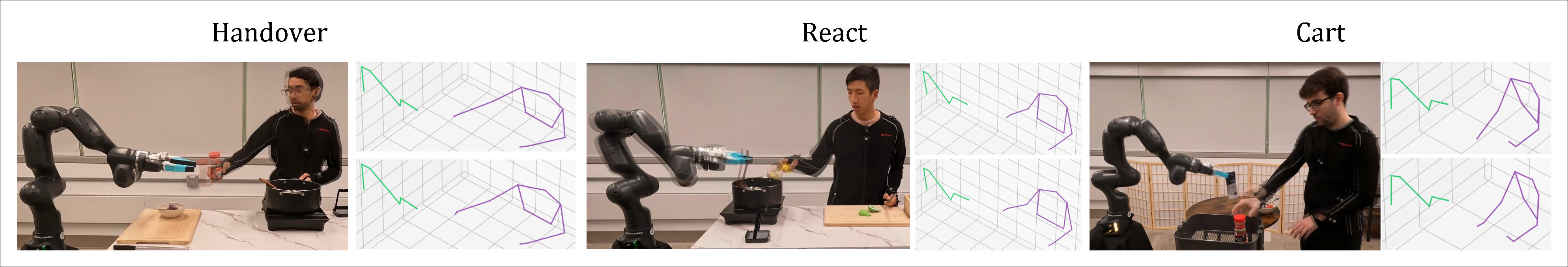}
\vspace{-2em}
\caption{Illustrative examples from the CoMaD dataset showcase its varied tasks and scenes for HRC in the kitchen. To visualize motion, we stack the corresponding RGB images.}
\vspace{-1.2em}
\label{CoMaD_hard_case}
\end{figure}

\begin{figure}[!t]
\centering
\includegraphics[width=3.5in]{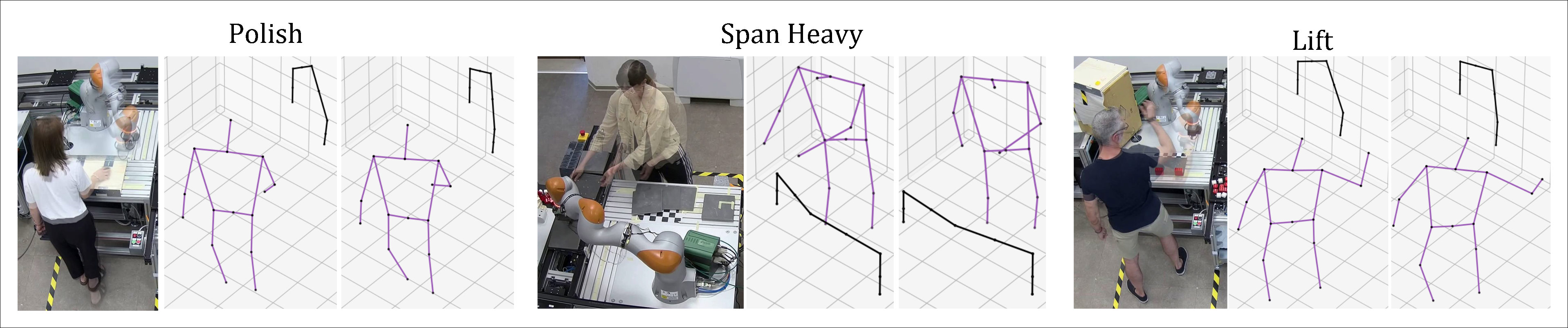}
\vspace{-2em}
\caption{Illustrative examples from the CHICO dataset show HRC scenarios in real-world factory assembly lines.}
\vspace{-1.5em}
\label{CHICO_hard_case}
\end{figure}

\begin{figure}[!t]
\centering
\includegraphics[width=3.5in]{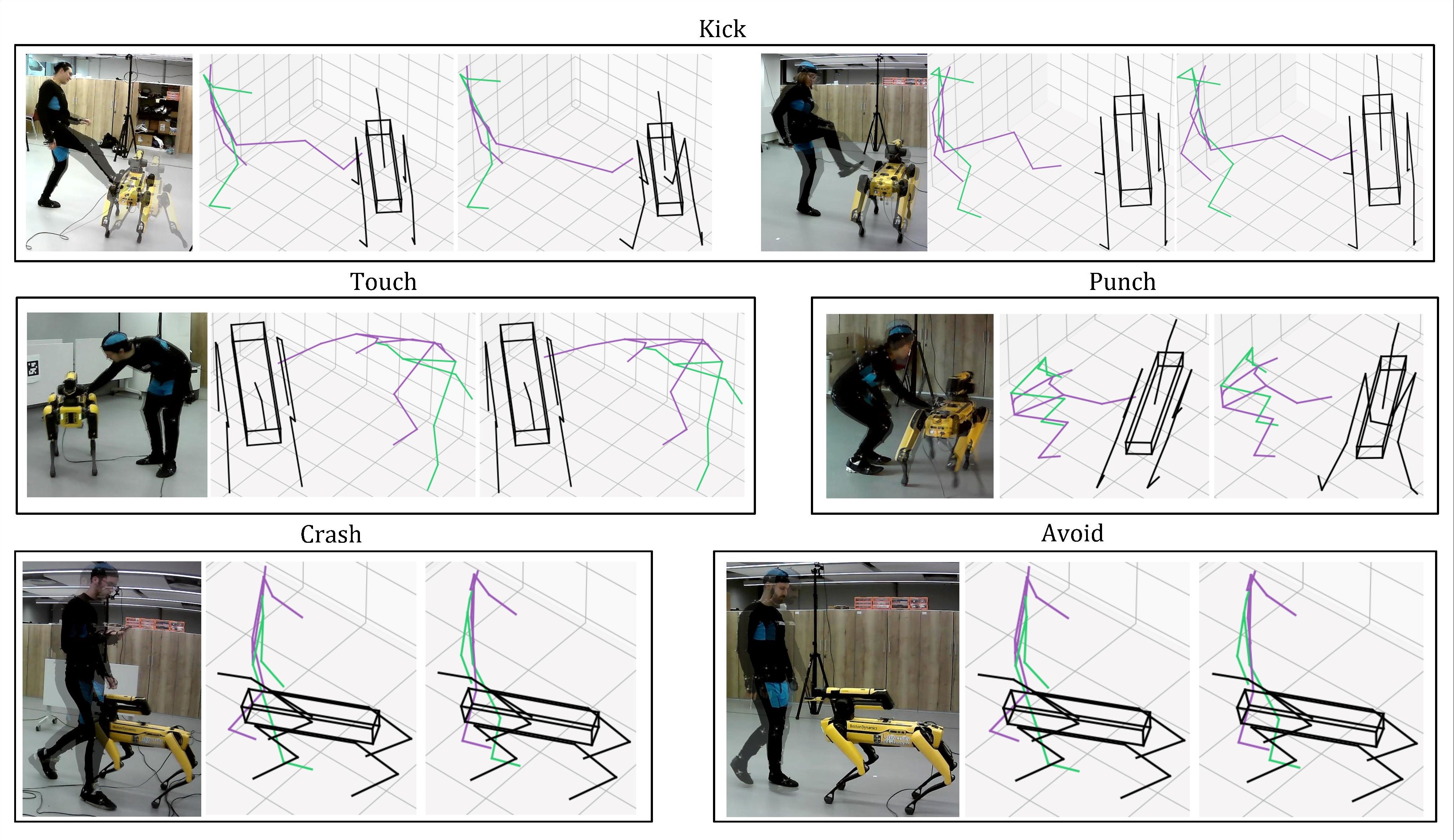}
\vspace{-2em}
\caption{Illustrative examples from the HARPER dataset showcase its diverse poses and rich human-robot interactions. }
\vspace{-1.2em}
\label{HARPER_hard_case}
\end{figure}

\textbf{Collaborative Manipulation Dataset (CoMaD)} \cite{kedia2024interact} 
The CoMaD dataset captures human-robot collaboration in a kitchen setting across five long-horizon tasks. It contains data from over 14 unique users interacting with a robotic arm that performs human-like assistive actions, such as handing over items, stirring food, and transferring objects from a cart, as depicted in Fig.~\ref{CoMaD_hard_case}. The dataset provides 3D skeleton annotations for both the human upper body and the robotic arm, with each human pose consisting of 11 joints and each robot pose comprising 10 joints.

\textbf{Cobots and Humans in Industrial Collaboration (CHICO)} \cite{sampieri2022pose} The CHICO dataset contains 3D pose sequences of 20 human operators collaborating with robotic arms in industrial assembly lines. Each human skeleton is represented by 15 joints, and each robot is represented by 9 joints. For our evaluation, we allocated 2 subjects for validation, 4 for testing, and the remaining 14 for training.
As shown in Fig.~\ref{CHICO_hard_case}, the robotic arm performs typical roles in CHICO scenarios, including alternating with humans to polish workbenches and collaborating side by side to complete concurrent tasks. 

\textbf{Human from an Articulated Robot Perspective (HARPER)} 
\cite{avogaro2024exploring} captures 3D dyadic interactions between humans and the Spot quadruped robot across 15 action categories, 10 of which involve physical contact. The dataset provides 3D skeleton annotations for both agents (21 human and 23 robot joints) and is split into 12 subjects for training and 5 for testing. As illustrated in Fig.~\ref{HARPER_hard_case}, HARPER features complex dynamics, including characteristic robot oscillations during contact, and realistically simulates collision-avoidance scenarios, making abrupt transitions such as walking to stopping especially difficult to predict without explicit contextual~cues.

\subsection{Implementation Details}
The model was trained on a single NVIDIA GeForce RTX 4090 GPU with a batch size of 256. Training required approximately 6 GB of memory and took approximately 2 hours to complete. For the HARPER dataset, training proceeded for 31,000 iterations, with an initial learning rate of 0.0003 for the first 30,000 iterations, which was then reduced to 0.00001. The model architecture processed 60 input frames to predict the subsequent 30 frames.

\begin{table*}[t]
\caption{Quantitative comparisons on CoMaD. ``H'' denotes human motion input, and ``HR'' denotes human and robot motion input. Our approach achieves SOTA performance at all time steps.}
\vspace{-1.5em}
\label{CoMaD-result}
\begin{center}
\renewcommand{\arraystretch}{1}
\setlength{\tabcolsep}{3mm}
\begin{tabular}{c|c| ccccc ccccc |c }
\toprule[1pt]
Method  & Input  & 100 & 200 & 300 &	400 &	500 &	600 & 700 &	800 & 900	& 1000 & Avg.
\\ \midrule 
SeS-GCN\cite{sampieri2022pose} & H & 21.79 & 28.00 & 33.68 & 39.15 & 44.06 & 48.70 & 53.05 & 57.06 & 60.78 & 64.10 & 45.04 \\
(ECCV'22) & HR & 17.34 & 23.18 & 29.08 & 35.07 & 40.84 & 46.15 & 50.99 & 55.55 & 59.75 & 63.48 & 42.14 \\ \midrule
{siMLPe}\cite{guo2023back} &  H & 6.01 	&	11.33 	&	16.63 	&	21.94 	&	27.09 	&	31.91 	&	36.43 	&	40.58 	&	44.34 	&	47.93 	&	28.42 \\
{(WACV'23)} &  HR & 5.91 	&	11.64 	&	17.51 	&	23.29 	&	28.63 	&	33.50 	&	38.09 	&	42.29 	&	46.18 	&	49.91 	&	29.69  \\ \midrule
PoseTron\cite{yasar2024posetron}  & HR & 10.71 & 18.87 & 26.95 & 32.66 & 38.69 & 43.08 & 47.48 & 50.91 & 54.50 & 57.34 & 38.12 \\ \midrule
{GCNext}\cite{wang2024gcnext} & H & 5.89 	&	11.33 	&	16.75 	&	22.10 	&	27.18 	&	31.81 	&	36.04 	&	39.99 	&	43.62 	&	47.00 	&	28.17  \\
 (AAAI'24)  & HR & 5.64 	&	11.00 	&	16.43 	&	21.85 	&	26.91 	&	31.61 	&	36.05 	&	40.10 	&	43.87 	&	47.53 	&	28.10 \\ \midrule
SiC\cite{wang2024skeleton} & H & 9.09 & 14.56 & 19.98 & 25.01 & 29.76 & 33.62 & 37.94 & 41.19 & 44.77 & 47.90 & 30.28 \\
(CVPR'24) & HR & 9.94 & 15.15 & 20.19 & 24.81 & 29.5 & 33.96 & 37.79 & 40.97 & 44.19 & 46.94 & 30.35 \\  \midrule
IAFormer\cite{xiao2024multi} & H & 7.58 & 13.69 & 19.38 & 24.76 & 30.16 & 35.02 & 39.36 & 43.20 & 46.56 & 49.66 & 30.94 \\
(ECCV'24) & HR & 8.57 & 15.43 & 22.00 & 28.19 & 34.45 & 40.36 & 45.83 & 50.97 & 55.88 & 60.37 & 36.21 \\ \midrule

CHAMP\cite{zhang2024champ}                                             & H     & 7.49          & 13.36          & 18.83          & 24.07          & 29.15          & 33.74          & 37.77          & 41.45          & 44.85          & 48.04          & 29.87          \\
(TCSVT'24)                                        & HR    & 7.87          & 13.99          & 19.40          & 24.83          & 29.79          & 34.44          & 38.61          & 42.43          & 45.90          & 49.23          & 30.65          \\ \hline
ALIEN \cite{wei2025alien}                                            & H     & 23.79         & 27.88          & 32.04          & 36.48          & 40.97          & 44.73          & 48.67          & 51.84          & 55.18          & 58.25          & 41.98          \\
(CVPR'25)                                         & HR    & 38.30         & 41.45          & 44.78          & 48.09          & 51.68          & 55.32          & 58.28          & 61.96          & 64.81          & 67.07          & 53.17          \\ \hline

\multirow{2}{*}{Ours} & H & 5.69 & 10.97 & 16.40 & 21.70 & 26.67 & 31.36 & 35.70 & 39.72 & 43.50 & 47.11 & 27.88 \\
 & HR & \textbf{5.50} & \textbf{10.60} & \textbf{15.76} & \textbf{20.92} & \textbf{25.88} & \textbf{30.54} & \textbf{34.94} & \textbf{39.13} & \textbf{43.01} & \textbf{46.70} & \textbf{27.30} \\ \bottomrule[1pt]
\end{tabular}
\end{center}
\vspace{-2em}
\end{table*}

\begin{table*}[t]
\caption{Quantitative comparisons on CHICO. Our approach achieves SOTA performance at all time steps.}  
\vspace{-1.5em}
\label{CHICO-result} 
\begin{center}
\setlength{\tabcolsep}{3.5mm}
\renewcommand{\arraystretch}{1}
\begin{tabular}{c|c| cccc cccc |c }  
\toprule[1pt]
Method	&	input	&	80	&	160	&	320	&	400	&	560	&	720	&	880	&	1000 & Avg.
\\ \midrule

SeS-GCN\cite{sampieri2022pose} & H & 16.72 & 28.11 & 46.75 & 53.58 & 64.30 & 72.71 & 80.40 & 86.04 & 56.08 \\
(ECCV'22) & HR & 18.03 & 30.34 & 50.65 & 57.91 & 68.92 & 77.54 & 85.51 & 91.30 & 60.03 \\ \midrule

siMLPe\cite{guo2023back}	&H&	13.51	&	24.04	&	40.95	&	47.58	&	58.62	&	68.40	&	77.10	&	83.03 & 51.65\\

{(WACV'23)}	&HR	& 13.23	& {23.41}	& {40.11}	& {46.78}	&	{57.33}	&	{66.86}	&	{75.05}	&	{80.18} & 50.37 \\
\midrule

PoseTron\cite{yasar2024posetron} & HR & 19.31 & 33.35 & 54.94 & 63.06 & 75.24 & 84.18 & 91.76 & 96.77 & 64.83 \\ \midrule

GCNext\cite{wang2024gcnext}	&	H	&	13.94	&	24.84	&	42.61	&	49.16	&	59.95	&	70.00	&	78.54	&	84.16 &52.90 \\
 
 {(AAAI'24)} &HR	&13.37	&	23.89	&	41.03	&	47.55	&	58.33	&	68.03	&	76.22	&	81.80  & 51.28\\
\midrule

SiC \cite{wang2024skeleton}& H & 19.10 & 29.49 & 48.57 & 56.60 & 68.35 & 79.71 & 89.32 & 94.46 & 60.70 \\
(CVPR'24) & HR & 18.00 & 27.68 & 46.11 & 52.73 & 62.84 & 71.89 & 79.47 & 84.09 & 55.35 \\
\midrule

IAFormer \cite{xiao2024multi} & H & 15.40 & 27.42 & 45.72 & 52.94 & 64.98 & 74.74 & 82.54 & 88.15 & 56.49 \\
(ECCV'24) & HR & 16.10 & 29.78 & 54.76 & 64.97 & 82.94 & 97.87 & 110.69 & 120.39 & 72.19 \\
\midrule

CHAMP \cite{zhang2024champ}                                            & H     & 14.40          & 25.68          & 43.30          & 49.97          & 61.44          & 70.71          & 78.27          & 83.54          & 53.41          \\
(TCSVT'24)                                        & HR    & 14.35          & 25.44          & 42.81          & 49.31          & 60.19          & 69.42          & 77.32          & 82.59          & 52.68          \\ \hline
ALIEN\cite{wei2025alien}                                             & H     & 21.14          & 30.92          & 45.69          & 51.17          & 60.34          & 68.61          & 75.71          & 80.40          & 54.25          \\
(CVPR'25)                                       & HR    & 21.98          & 31.58          & 45.69          & 50.99          & 60.06          & 68.00          & 74.78          & 79.62          & 54.09          \\ \hline

\multirow{2}{*}{Ours}	&  H & 13.23 & 23.42 & 40.04 & 46.62 & 57.22 & 66.91 & 75.48 & 81.01 & 50.49  \\ 

& HR & \textbf{13.06} & \textbf{23.05} & \textbf{39.45} & \textbf{45.79} & \textbf{55.93} & \textbf{65.40} & \textbf{73.43} & \textbf{78.69} & \textbf{49.35}  \\ \bottomrule[1pt]

\end{tabular}
\end{center}
\vspace{-2em}
\end{table*}

\subsection{Baseline}
We conduct extensive comparisons with state-of-the-art approaches across three categories: HMP during HRC, single-person motion prediction, and multi-person motion prediction.
In the context of HRC, we compare our method with two recent methods: SeS-GCN \cite{sampieri2022pose} and PoseTron \cite{yasar2024posetron}. Although both methods are evaluated on HRC datasets, they lack explicit modeling of human-robot interactions. Specifically, SeS-GCN can be regarded as a single-person method, whereas PoseTron can be viewed as a multi-person approach.
Additionally, we compare our work with state-of-the-art methods in single-person and multi-person motion prediction, including siMLPe \cite{guo2023back}, GCNext \cite{wang2024gcnext}, SiC \cite{wang2024skeleton},  IAFormer \cite{xiao2024multi}, CHAMP \cite{zhang2024champ} and ALIEN \cite{wei2025alien}. Notably, IAFormer serves as both a multi-person and single-person motion prediction baseline, while the others are designed for single-person motion prediction.

To ensure a fair comparison, we evaluate all methods under two settings: (1) using only human motion as input, and (2) using both human and robot motion as input. For single-person motion prediction methods, we treat the human and robot as a unified entity in the supplementary experiments. For our robot-aware HMP framework, we also test a simplified version of our model without the LGSI module and AWA, using only human motion as input.

For multi-person motion prediction methods such as PoseTron~\cite{yasar2024posetron} and IAFormer~\cite{xiao2024multi}, which require all agents to have the same number of joints, we address this discrepancy by duplicating the root joint for agents with fewer joints. This ensures compatibility with the evaluation framework while maintaining the structural integrity of the motion data.

\subsection{Quantitative results}
\textbf{Comparison of the CoMaD and CHICO Dataset Results}
In Table~\ref{CoMaD-result} and Table~\ref{CHICO-result}, we present the results derived from the CoMaD and CHICO datasets, using the MPJPE as the metric, at different time steps across all actions. Despite the strong results achieved by the baseline methods, our method improves prediction accuracy. 
Interestingly, when the same method is used, a model that incorporates human-robot (HR) motion input does not consistently outperform a model that only uses human (H) motion input. 
The reason is that without effective modeling of interaction dynamics, the additional global information provided by the HR input fails to increase prediction accuracy and may even degrade performance due to interference.

\begin{table*}[t]
\caption{Quantitative comparisons on HARPER. Ours-HR demonstrates a marginal advantage over the baselines.}
\vspace{-1.5em}
\label{HARPER-result}
\begin{center}
\setlength{\tabcolsep}{3mm}
\begin{tabular}{c|c| ccccc ccccc |c}
\toprule[1pt]
Method  & Input  & 100 & 200 & 300 &	400 &	500 &	600 & 700 &	800 & 900	& 1000 & Avg.
\\ \midrule
SeS-GCN\cite{sampieri2022pose} & H & 42.48 & 49.92 & 59.80 & 68.30 & 79.79 & 93.41 & 105.13 & 119.13 & 133.21 & 148.80 & 90.00 \\
(ECCV'22) & HR & 48.10 & 55.76 & 64.10 & 76.66 & 86.56 & 98.59 & 113.31 & 125.96 & 139.10 & 154.16 & 96.23 \\ \hline

 {siMLPe}\cite{guo2023back} &  H & 5.81 	&	13.47 	&	22.31 	&	32.38 	&	43.17 	&	54.46 	&	66.49 	&	79.23 	&	92.51 	&	106.16 	&	51.60 \\
{(WACV'23)} &  HR & 5.93 	&	13.61 	&	22.65 	&	32.85 	&	43.58 	&	54.81 	&	66.78 	&	79.32 	&	92.27 	&	105.49 	&	51.73 \\ \hline

PoseTron\cite{yasar2024posetron} & HR & 19.87 & 36.91 & 53.12 & 69.42 & 84.82 & 100.50 & 115.96 & 131.06 & 145.87 & 160.16 & 91.77 \\ \hline

{GCNext}\cite{wang2024gcnext} & H & 5.70 	&	13.77 	&	23.28 	&	33.53 	&	44.04 	&	55.08 	&	66.95 	&	79.45 	&	92.40 	&	105.65 	&	51.98  \\
 (AAAI'24)  & HR & 6.02 	&	13.85 	&	23.13 	&	33.56 	&	44.40 	&	55.73 	&	67.81 	&	80.28 	&	92.97 	&	105.96 	&	52.37  \\  \hline

SiC\cite{wang2024skeleton} & H & 14.29 & 25.37 & 37.63 & 49.9 & 62.63 & 75.86 & 89.72 & 103.99 & 118.86 & 134.29 & 71.25 \\
(CVPR'24) & HR & 21.06 & 31.67 & 43.85 & 56.91 & 70.55 & 84.46 & 98.26 & 112.59 & 126.49 & 141.47 & 78.73 \\ \hline

IAFormer\cite{xiao2024multi} & H & 9.88 & 19.82 & 30.13 & 40.92 & 52.22 & 63.97 & 76.25 & 88.94 & 101.79 & 114.83 & 59.88 \\
(ECCV'24) & HR & 13.44 & 25.63 & 38.34 & 51.74 & 65.54 & 80.34 & 95.78 & 112.17 & 128.49 & 144.98 & 75.64 \\ \hline
CHAMP\cite{zhang2024champ}                                             & H     & 8.78          & 17.36          & 26.46          & 36.31          & 46.85          & 57.88          & 69.51          & 81.86          & 94.87          & 108.16          & 54.80          \\
(TCSVT'24)                                        & HR    & 9.06          & 17.56          & 27.02          & 37.31          & 48.22          & 59.69          & 71.55          & 83.46          & 95.59          & 107.73          & 55.72          \\ \hline
ALIEN\cite{wei2025alien}                                             & H     & 30.06         & 39.98          & 51.46          & 62.65          & 73.30          & 84.87          & 96.28          & 109.01         & 122.29         & 136.15          & 80.61          \\
(CVPR'25)                                         & HR    & 42.66         & 50.58          & 61.58          & 72.31          & 83.48          & 94.88          & 106.26         & 117.87         & 129.41         & 141.07          & 90.01          \\ \hline

\multirow{2}{*}{Ours} & H & 5.13 & 12.15 & 20.65 & 30.39 & 40.76 & 51.90 & 64.02 & 76.81 & 90.07 & 103.69 & 49.56 \\

 & HR & \textbf{4.94} & \textbf{11.71} & \textbf{19.86} & \textbf{29.25} & \textbf{39.30} & \textbf{49.97} & \textbf{61.48} & \textbf{73.81} & \textbf{86.72} & \textbf{100.11} & \textbf{47.71} \\ \bottomrule[1pt]

\end{tabular}
\end{center}
\vspace{-2em}
\end{table*}

\begin{table*}[!t]
\caption{Action-wise results on HARPER. Our method significantly outperforms the baselines across various actions.}  
\vspace{-1em}
\label{HARPER-actionwise}
\begin{center}  
\setlength{\tabcolsep}{1.5mm}
\renewcommand{\arraystretch}{1.07}
\begin{tabular}{c|ccccc|ccccc|ccccc}
\toprule[1pt]
Action & \multicolumn{5}{c|}{Walk + Crash Sideways} & \multicolumn{5}{c|}{Walk + Crash Backwards} & \multicolumn{5}{c}{Walk + Avoid Right} \\ \hline
milliseconds & 0.1s & 0.4s & 0.7s & 1s & Avg. & 0.1s & 0.4s & 0.7s & 1s & Avg. & 0.1s & 0.4s & 0.7s & 1s & Avg. \\ \hline
SeS-GCN-H\cite{sampieri2022pose} & 20.14 & 73.76 & 139.67 & 216.66 & 98.93 & 92.74 & 132.34 & 182.45 & 243.55 & 160.56 & 83.76 & 95.76 & 119.93 & 154.39 & 111.45 \\
SeS-GCN-HR\cite{sampieri2022pose} & 56.98 & 105.36 & 167.75 & 239.79 & 139.78 & 53.63 & 101.88 & 162.11 & 231.36 & 134.81 & 46.31 & 67.08 & 98.34 & 133.86 & 84.59 \\ \hline
siMLPe-H\cite{guo2023back} & 9.87 & 58.68 & 126.81 & 212.97 & 98.87 & 11.49 & 66.49 & 144.94 & 240.29 & 112.53 & 5.18 & 29.08 & 59.62 & 94.83 & 46.15 \\
siMLPe-HR\cite{guo2023back} & 9.76 & 55.19 & 116.77 & 191.98 & 90.95 & 10.93 & 61.65 & 130.57 & 204.26 & 99.95 & 4.97 & 27.40 & 57.83 & 96.07 & 45.20 \\ \hline

PoseTron-HR\cite{yasar2024posetron} & 31.34 & 112.05 & 184.62 & 263.74 & 148.09 & 34.68 & 130.66 & 220.78 & 307.96 & 174.37 & 23.17 & 78.89 & 134.38 & 188.42 & 106.23 \\ \hline
GCNext-H\cite{wang2024gcnext} & 9.32 & 52.74 & 112.02 & 186.26 & 87.44 & 10.44 & 60.86 & 131.11 & 217.73 & 102.15 & 4.55 & 26.76 & 58.04 & 96.22 & 45.08 \\
GCNext-HR\cite{wang2024gcnext} & 10.06 & 58.30 & 124.63 & 207.92 & 97.28 & 11.22 & 63.78 & 134.19 & 213.03 & 103.45 & 5.15 & 28.14 & 59.65 & 95.78 & 46.03 \\ \hline

SiC-H\cite{wang2024skeleton} & 20.14 & 73.76 & 139.67 & 216.66 & 98.93 & 21.65 & 84.47 & 159.14 & 238.29 & 111.73 & 12.43 & 47.38 & 96.46 & 157.07 & 67.14 \\
SiC-HR\cite{wang2024skeleton} & 23.54 & 87.99 & 162.84 & 248.63 & 128.93 & 25.95 & 88.09 & 151.39 & 212.97 & 119.46 & 16.94 & 55.27 & 107.56 & 170.79 & 85.68 \\ \hline
IAFormer-H\cite{xiao2024multi} & 18.02 & 70.40 & 140.33 & 219.00 & 109.23 & 18.89 & 76.98 & 145.25 & 229.03 & 114.70 & 9.89 & 36.16 & 69.75 & 112.64 & 55.83 \\
IAFormer-HR\cite{xiao2024multi} & 19.27 & 75.97 & 144.16 & 224.69 & 114.03 & 20.48 & 79.74 & 150.66 & 233.13 & 119.13 & 11.92 & 43.29 & 82.68 & 130.67 & 65.89 \\ \hline
\multicolumn{1}{c|}{CHAMP-H\cite{zhang2024champ}}                                              & 12.86         & 58.37          & 124.55          & 211.85          & \multicolumn{1}{c|}{98.58}          & 14.15         & 65.75          & 150.13          & 258.09          & \multicolumn{1}{c|}{117.58}         & 7.16          & 28.83          & 58.97          & 92.17          & 45.82                     \\
\multicolumn{1}{c|}{CHAMP-HR\cite{zhang2024champ}}                                             & 13.21         & 60.76          & 128.87          & 213.15          & \multicolumn{1}{c|}{101.01}         & 14.21         & 67.74          & 140.56          & 223.85          & \multicolumn{1}{c|}{109.21}         & 7.46          & 31.39          & 66.52          & 103.96         & 51.18                     \\ \hline
\multicolumn{1}{c|}{ALIEN-H\cite{wei2025alien}}                                              & 37.61         & 88.39          & 148.17          & 228.51          & \multicolumn{1}{c|}{122.99}         & 38.17         & 95.29          & 162.27          & 236.97          & \multicolumn{1}{c|}{131.38}         & 26.59         & 58.41          & 89.75          & 130.25         & 75.53                     \\
\multicolumn{1}{c|}{ALIEN-HR\cite{wei2025alien}}                                             & 53.32         & 105.46         & 166.74          & 233.56          & \multicolumn{1}{c|}{138.39}         & 48.60         & 105.71         & 168.20          & 221.80          & \multicolumn{1}{c|}{136.27}         & 42.43         & 72.61          & 110.27         & 146.57         & 92.46                     \\ \hline

Ours-H & 8.54 & 49.37 & 109.56 & 193.24 & 86.57 & 9.59 & 57.55 & 129.24 & 224.48 & 101.23 & 4.03 & 21.67 & 47.79 & 79.61 & 37.08 \\
Ours-HR & \textbf{8.40} & \textbf{47.76} & \textbf{104.90} & \textbf{183.40} & \textbf{82.89} & \textbf{9.49} & \textbf{58.52} & \textbf{124.36} & \textbf{209.09} & \textbf{97.21} & \textbf{4.00} & \textbf{20.04} & \textbf{43.97} & \textbf{75.94} & \textbf{34.72} \\ \hline
Action & \multicolumn{5}{c|}{Circular Follow + Touch} & \multicolumn{5}{c|}{Circular Follow + Avoid} & \multicolumn{5}{c}{Circular Follow + Crash} \\ \hline
milliseconds & 0.1s & 0.4s & 0.7s & 1s & Avg. & 0.1s & 0.4s & 0.7s & 1s & Avg. & 0.1s & 0.4s & 0.7s & 1s & Avg. \\ \hline
SeS-GCN-H\cite{sampieri2022pose} & 87.01 & 94.00 & 115.54 & 148.35 & 110.06 & 65.65 & 69.35 & 75.48 & 87.72 & 74.53 & 74.03 & 86.49 & 109.25 & 143.32 & 101.26 \\
SeS-GCN-HR\cite{sampieri2022pose} & 43.78 & 72.11 & 103.14 & 144.48 & 89.85 & 36.51 & 52.45 & 61.73 & 83.19 & 58.34 & 42.23 & 69.53 & 98.27 & 139.31 & 86.30 \\ \hline
siMLPe-H\cite{guo2023back} & 4.45 & 27.95 & 58.26 & 92.10 & 44.75 & 2.87 & 14.50 & 28.48 & 44.39 & 22.19 & 4.87 & 27.72 & 56.27 & 89.77 & 43.76 \\
siMLPe-HR\cite{guo2023back} & 4.80 & 27.59 & 57.11 & 89.89 & 43.96 & 3.32 & 15.20 & 29.37 & 45.21 & 22.93 & 5.11 & 27.57 & 55.53 & 84.32 & 42.57 \\ \hline

PoseTron-HR\cite{yasar2024posetron} & 18.06 & 68.03 & 115.94 & 163.56 & 91.47 & 10.23 & 31.86 & 53.47 & 77.70 & 43.03 & 16.13 & 59.42 & 106.38 & 153.34 & 83.51 \\ \hline
GCNext-H\cite{wang2024gcnext} & 4.17 & 26.96 & 56.97 & 92.25 & 43.99 & 2.71 & 13.68 & 27.12 & 42.27 & 21.09 & 4.51 & 25.95 & 52.81 & 84.48 & 41.09 \\
GCNext-HR\cite{wang2024gcnext} & 4.79 & 27.75 & 56.78 & 88.14 & 43.60 & 3.42 & 15.70 & 30.71 & 47.29 & 23.91 & 5.19 & 27.91 & 55.97 & 85.33 & 43.01 \\ \hline

SiC-H\cite{wang2024skeleton} & 11.55 & 40.01 & 71.92 & 109.84 & 51.59 & 8.17 & 23.42 & 39.81 & 59.00 & 29.26 & 12.16 & 40.20 & 71.52 & 105.40 & 51.41 \\
SiC-HR\cite{wang2024skeleton} & 13.90 & 44.78 & 78.77 & 113.73 & 62.36 & 9.85 & 29.60 & 52.00 & 74.70 & 41.27 & 13.64 & 41.79 & 69.45 & 97.47 & 55.50 \\ \hline
IAFormer-H\cite{xiao2024multi} & 9.24 & 37.07 & 68.00 & 103.24 & 53.73 & 6.02 & 19.41 & 32.83 & 47.15 & 26.33 & 9.37 & 36.46 & 65.34 & 96.41 & 51.56 \\
IAFormer-HR\cite{xiao2024multi} & 11.00 & 43.43 & 82.84 & 128.01 & 65.41 & 7.26 & 27.84 & 53.56 & 82.85 & 42.18 & 11.14 & 43.67 & 81.67 & 123.92 & 64.33 \\ \hline
\multicolumn{1}{c|}{CHAMP-H\cite{zhang2024champ}}                                              & 6.11          & 28.30          & 57.64           & 93.03           & \multicolumn{1}{c|}{45.14}          & 4.55          & 16.25          & 31.13           & 49.03           & \multicolumn{1}{c|}{24.78}          & 6.75          & 28.44          & 55.35          & 88.61          & 43.83                     \\
\multicolumn{1}{c|}{CHAMP-HR\cite{zhang2024champ}}                                             & 7.54          & 31.42          & 62.25           & 92.81           & \multicolumn{1}{c|}{47.90}          & 6.03          & 19.15          & 35.76           & 54.95           & \multicolumn{1}{c|}{28.46}          & 7.82          & 30.24          & 55.47          & 84.22          & 43.86                     \\ \hline
\multicolumn{1}{c|}{ALIEN-H\cite{wei2025alien}}                                              & 27.29         & 52.96          & 85.10           & 124.47          & \multicolumn{1}{c|}{71.34}          & 22.62         & 37.60          & 48.90           & 63.79           & \multicolumn{1}{c|}{43.17}          & 26.44         & 51.72          & 76.78          & 106.79         & 64.94                     \\
\multicolumn{1}{c|}{ALIEN-HR\cite{wei2025alien}}                                             & 40.26         & 65.39          & 97.22           & 131.98          & \multicolumn{1}{c|}{82.89}          & 33.50         & 47.94          & 61.96           & 74.10           & \multicolumn{1}{c|}{54.44}          & 37.83         & 59.73          & 86.46          & 114.18         & 73.82                     \\ \hline

Ours-H & 3.76 & 24.01 & 51.89 & 85.50 & 40.13 & 2.58 & 12.40 & 25.67 & 42.71 & 20.23 & 4.12 & 23.82 & 50.52 & 82.23 & 39.17 \\
Ours-HR & \textbf{3.58} & \textbf{23.27} & \textbf{50.65} & \textbf{82.69} & \textbf{38.97} & \textbf{2.52} & \textbf{12.05} & \textbf{24.37} & \textbf{39.88} & \textbf{19.21} & {\textbf{3.97}} & {\textbf{22.89}} & {\textbf{49.01}} & {\textbf{79.95}} & {\textbf{37.95}} \\

\bottomrule[1pt]
\end{tabular}
\end{center}
\vspace{-2em}
\end{table*}

\begin{table}[!t]
\centering
\caption{ {Efficiency comparison across hardware platforms. FLOPs/Params are theoretical; FPS is measured on an RTX 3060 and Jetson Nano.}}
\vspace{-1em}
\label{inference_time}
\small
\setlength{\tabcolsep}{1mm}
\renewcommand{\arraystretch}{1.2}
\begin{tabular}{l c c c cc}
\toprule
\multirow{2.5}{*}{Model} 
& \multirow{2.5}{*}{\begin{tabular}[c]{@{}c@{}}FLOPs\\ (M)$\downarrow$\end{tabular}} 
& \multirow{2.5}{*}{\begin{tabular}[c]{@{}c@{}}Para.\\ (M)$\downarrow$\end{tabular}} 
& \multirow{2.5}{*}{\begin{tabular}[c]{@{}c@{}}MPJPE\\ (mm)$\downarrow$\end{tabular}} 
& \multicolumn{2}{c}{Inference FPS $\uparrow$} \\ 
\cmidrule(lr){5-6}
 &  &  &  & RTX 3060 & {Jetson Nano} \\ 
\midrule
siMLPe-HR & 24.90 & \textbf{0.21} & 106.16 & \textbf{278.89} & {\textbf{37.08}} \\
GCNext-HR & 24.90 & \textbf{0.21} & 105.96 & 56.23 & {8.52} \\ 
\rowcolor{gray!10}
\textbf{Ours-HR} & \textbf{25.58} & 0.41 & \textbf{100.10} & 148.59 & {30.69} \\ 
\bottomrule
\end{tabular}
\vspace{-1.5em}
\end{table}

\textbf{Comparison of the HARPER Dataset Results}
The results derived from the HARPER dataset are shown in Table~\ref{HARPER-result}. Our model notably outperforms the baseline models at every time step in both the Ours-HR and Ours-H configurations.

The action-wise results for six exemplary actions derived from the HARPER dataset are displayed in Table~\ref{HARPER-actionwise}. Our approach consistently outperforms the baselines across various actions.
For actions that require precise interaction prediction, particularly ``Walk + Crash'' and ``Circular Follow + Touch'' (both of which demand accurate forecasting of action transitions and robot contact points), our method's superior performance demonstrates the essential role of proper human-robot interaction modeling.
{This trend further suggests that the action-wise error variance is closely related to motion complexity. In particular, actions involving abrupt collision, contact transition, or strong proximity constraints exhibit higher stochasticity and are therefore more difficult for deterministic predictors, which also explains the larger errors observed in crash-related categories.}

Notably, the CoMaD, CHICO, and HARPER datasets exhibit significant differences in aspects such as scenarios, robot types, and action characteristics. These differences highlight the broad applicability of our method across diverse settings.

When we compared the improvements in our model relative to the baseline methods across the three datasets, we observed that the greatest enhancement was achieved on the HARPER dataset. {This may be attributed to the fact that HARPER contains more skeletal joint points and a more diverse set of interaction categories, making it more challenging. In contrast, the CoMaD and CHICO datasets are relatively simpler, and thus, the siMLPe-HR and GCNext-HR models achieve suboptimal results by indirectly modeling interaction relationships.}

\textbf{Comparison of Computational Efficiency}
A comparative analysis of computational efficiency between our method and state-of-the-art approaches on the HARPER dataset for 1-second human motion prediction is presented in Table \ref{inference_time}. {To evaluate both theoretical efficiency and practical deployment capabilities, the experiments were conducted on both an NVIDIA GeForce RTX 3060 GPU and a compute-constrained edge device, the NVIDIA Jetson Nano. }
For models that incorporate human-robot (HR) motion input, our \textbf{Ours-HR} approach not only achieves state-of-the-art performance (MPJPE of 100.10 mm) but also demonstrates exceptional computational efficiency, fully satisfying the stringent real-time requirements for real-world and mobile robotic applications.

\subsection{Qualitative results}
\begin{figure}[!tp]
\centering
\includegraphics[width=3.5in]{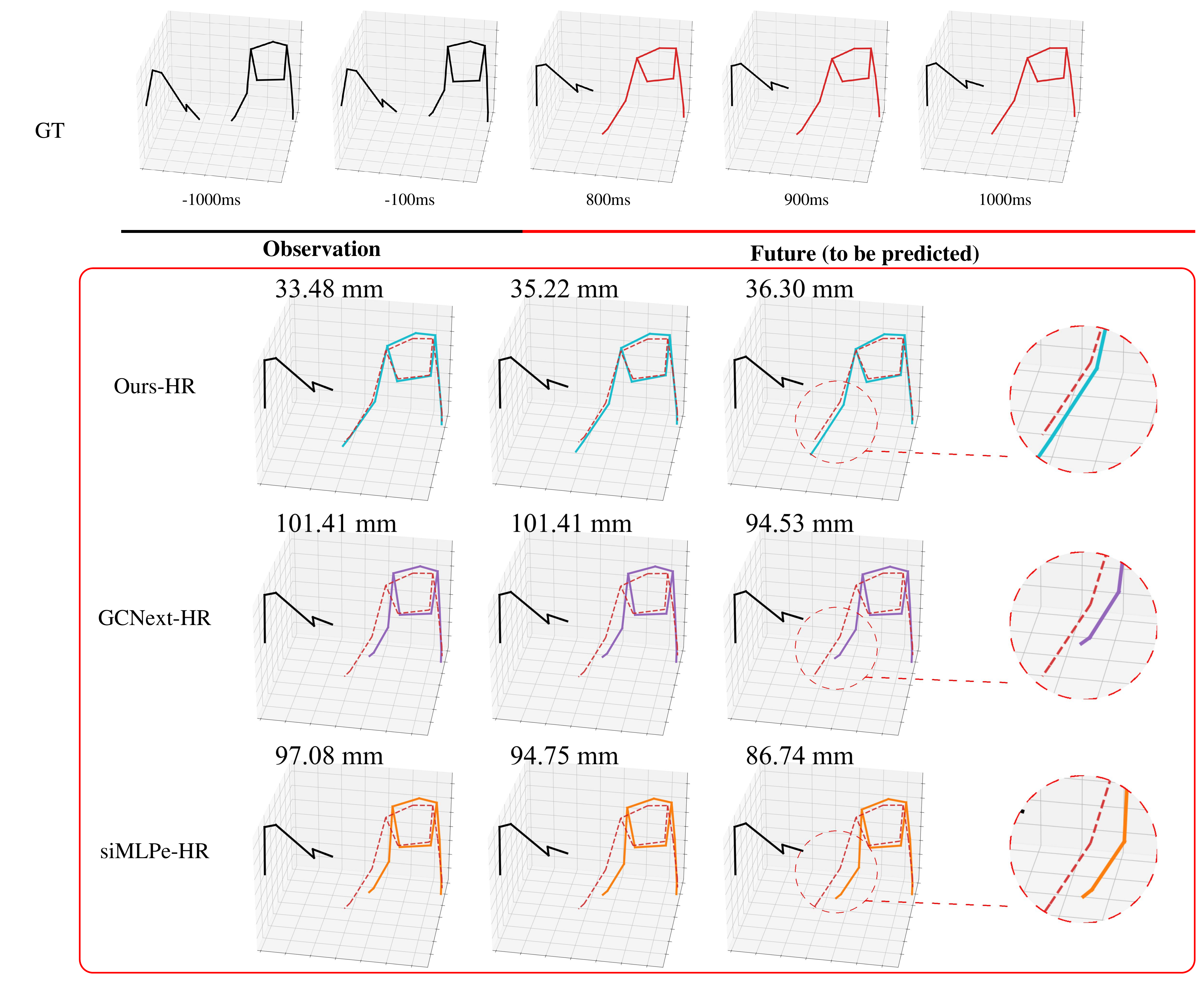}
\vspace{-2em}
\caption{
{
Qualitative comparison on the tabletop collaboration setting of CoMaD~\cite{kedia2024interact} between our method and prior approaches, including siMLPe~\cite{guo2023back} and GCNext~\cite{wang2024gcnext}. Robot motions are shown as ground truth to illustrate the human-robot interaction context. MPJPE values (lower is better) are reported above each prediction, and dashed circles highlight notable differences.}
}
\label{CoMaD_quali}
\vspace{-1em}
\end{figure}

A collaborative cooking scenario in a kitchen from the CoMaD dataset, where a human and a robotic arm sequentially reach adjacent bottles, is depicted in Fig.~\ref{CoMaD_quali}. In this HRC scene, both agents must observe each other's actions to plan subsequent movements effectively. Consequently, capturing the complete human-robot interaction context enables more accurate prediction of future motions. Compared with baseline methods, our approach achieves remarkably more accurate human pose predictions, particularly during object grasping. In contrast, other models fail to predict the human's intention to grasp the bottle, resulting in higher MPJPE values. 

\begin{figure}[!tp]
\centering
\includegraphics[width=3.5in]{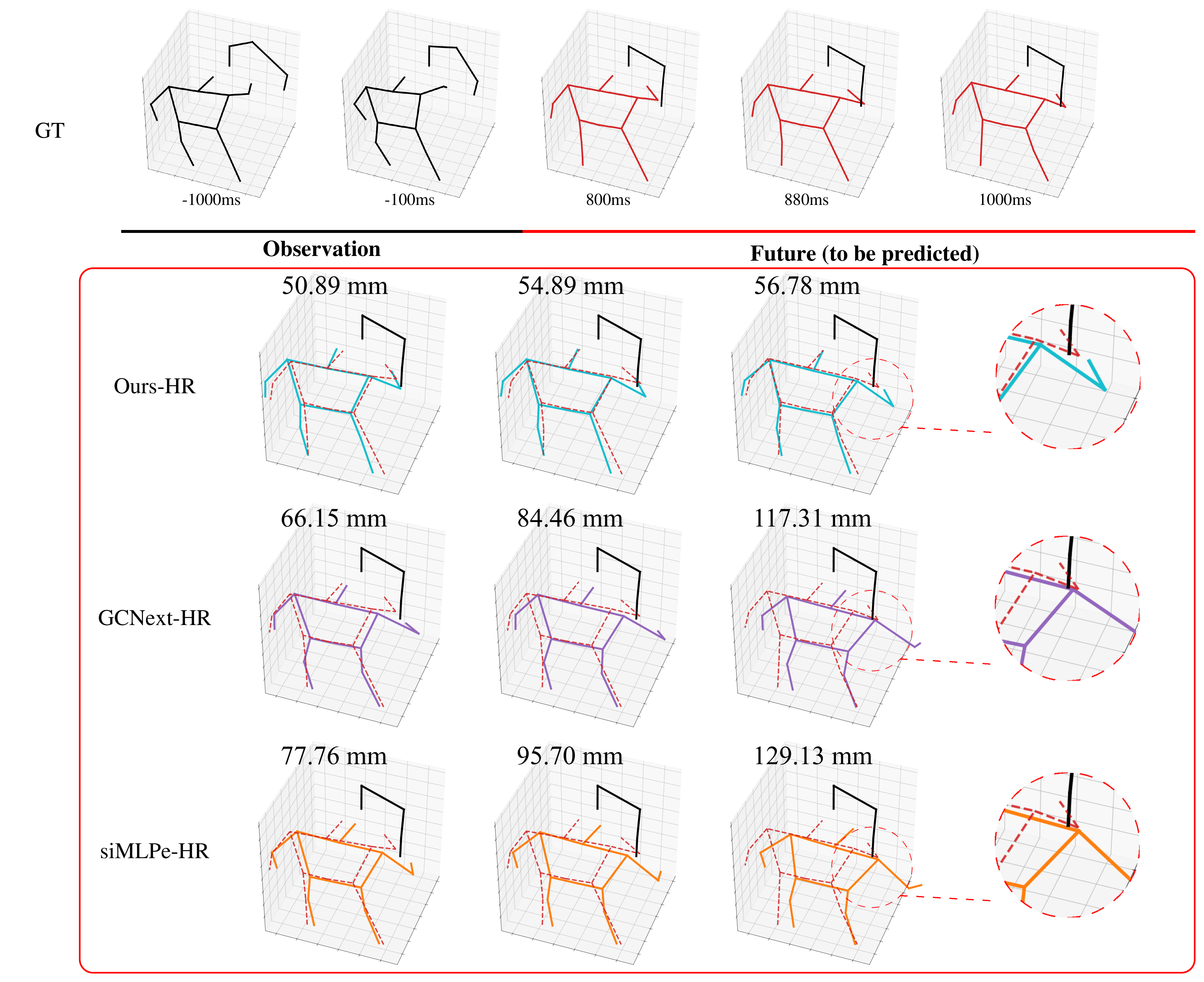}
\vspace{-2em}
\caption{
{
Qualitative comparison on the HRC scene of CHICO~\cite{sampieri2022pose} between our method and prior approaches, including siMLPe~\cite{guo2023back} and GCNext~\cite{wang2024gcnext}. 
}
}

\label{CHICO_quali}
\vspace{-1em}
\end{figure}

A representative sequence from the CHICO dataset is shown in Fig.~\ref{CHICO_quali}, highlighting the superior motion prediction capabilities of our proposed method. For relay-style human-robot object lifting tasks, our approach produces more plausible sequences, with accurate object placement and coordinated limb movements.
The motion of the robotic arm provides essential cues about target locations as humans move lifted objects upward. This demonstrates our interaction-aware method's enhanced understanding of the environmental context and human intentions, yielding more realistic motion sequences than the alternatives.

\begin{figure}[!tp]
\centering
\includegraphics[width=3.5in]{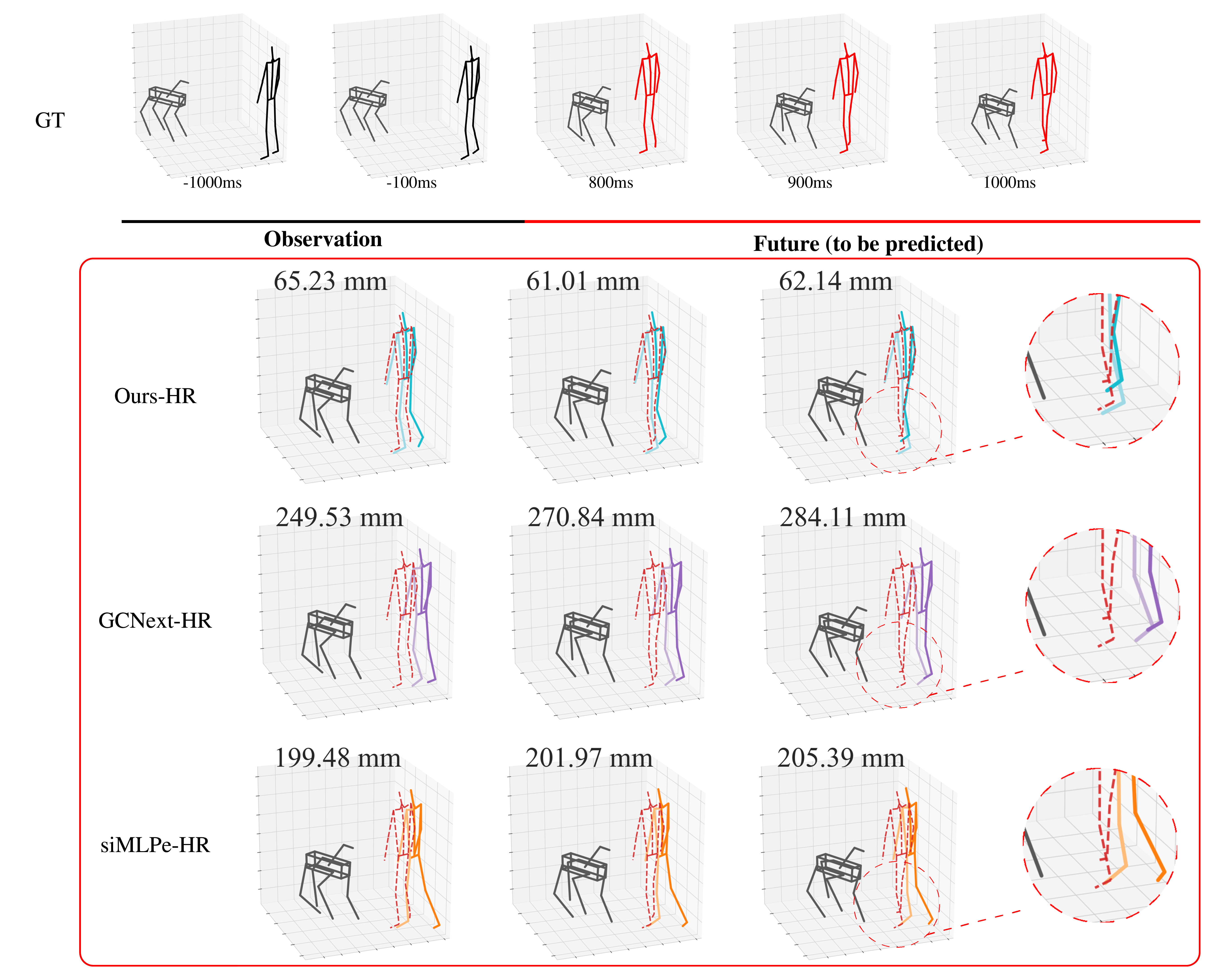}
\vspace{-2em}

\caption{
{Qualitative comparison on the ``Walk+Avoid right'' action of HARPER~\cite{avogaro2024exploring} between our method and prior approaches, including siMLPe~\cite{guo2023back} and GCNext~\cite{wang2024gcnext}. 
}
}
\label{touch_fig}
\end{figure}

A sample from the ``Walk+Avoid right'' action in the HARPER dataset, where a human and robot approach each other until the human performs collision avoidance by stopping and moving rightward, is shown in Fig.~\ref{touch_fig}. 
This interaction-intensive task necessitates modeling of mutual human-robot influences to accurately predict both the timing of action transitions and the corresponding human poses. 
In this sample, 
the Ours-HR method achieves significantly lower MPJPE values than all baseline methods do. These results validate the importance of our interaction-aware modeling approach for accurately predicting human motion in HRI scenarios.

\begin{table}[!t]
\caption{Ablation studies on Core Components of Model Structures. Experiments are Conducted With the HARPER Dataset. }  
\label{Ablation-arch}
\vspace{-1em}
\begin{center}
\setlength{\tabcolsep}{3.2mm}
\renewcommand{\arraystretch}{1.2}
\begin{tabular}{ccccc}
\toprule[1pt]
\multicolumn{1}{c|}{\multirow{2}{*}{Method}} & \multicolumn{4}{c}{MPJPE(mm)} \\ \cline{2-5} 
\multicolumn{1}{c|}{} & \multicolumn{1}{c}{100} & \multicolumn{1}{c|}{1000} & \multicolumn{1}{c|}{Avg.} & \multicolumn{1}{c}{Avg. $\triangle$} \\ \hline
\multicolumn{1}{c|}{ours-HR} & \multicolumn{1}{c}{\textbf{4.94}} & \multicolumn{1}{c|}{\textbf{100.11}} & \multicolumn{1}{c|}{\textbf{47.71}} & \multicolumn{1}{c}{-} \\ \hline
\multicolumn{1}{c|}{w/o LSI} & \multicolumn{1}{c}{4.97} & \multicolumn{1}{c|}{101.06} & \multicolumn{1}{c|}{48.12} & \multicolumn{1}{c}{0.40} \\ 
\multicolumn{1}{c|}{w/o GSI} & \multicolumn{1}{c}{5.14} & \multicolumn{1}{c|}{101.08} & \multicolumn{1}{c|}{48.69} & \multicolumn{1}{c}{0.97} \\ 
\multicolumn{1}{c|}{w/o LGSI} & 5.01 & \multicolumn{1}{c|}{101.82} & \multicolumn{1}{c|}{48.15} & \multicolumn{1}{c}{0.44} \\ \hline
\multicolumn{1}{c|}{w/o AWA} & \multicolumn{1}{c}{5.11} & \multicolumn{1}{c|}{102.62} & \multicolumn{1}{c|}{49.28} & \multicolumn{1}{c}{1.57} \\ \hline
\multicolumn{1}{c|}{w/o Traj Module} & \multicolumn{1}{c}{5.00} & \multicolumn{1}{c|}{102.09} & \multicolumn{1}{c|}{48.22} & \multicolumn{1}{c}{0.51} \\ 
\multicolumn{1}{c|}{w/o Pose Module} & 5.35 & \multicolumn{1}{c|}{109.16} & \multicolumn{1}{c|}{51.91} & 4.20 \\  
\bottomrule[1pt]

\end{tabular}
\end{center}
\vspace{-2em}
\end{table}

\begin{table}[t]
\caption{Ablation study on the implementation of the AWA. Comparison of the performance of the full model (Ours-HR) with variations in which specific AWA components (LN, truncation, and sumNorm) and context are excluded. ``w/o context'' refers to using learnable parameters instead.
}
\begin{center}
\label{Ablation-AWI}
\vspace{-0.5em}
\setlength{\tabcolsep}{3mm}
\renewcommand{\arraystretch}{1.2}
\begin{tabular}{c|ccccc}
\toprule[1pt]
\multirow{2}{*}{Method} & \multicolumn{5}{c}{MPJPE(mm)} \\ \cline{2-6} 
 & \multicolumn{1}{c}{100} & \multicolumn{1}{c}{400} & \multicolumn{1}{c}{700} & \multicolumn{1}{c|}{1000} & Avg. \\ \hline
Ours-HR & \multicolumn{1}{c}{\textbf{4.94}} & \multicolumn{1}{c}{\textbf{29.25}} & \multicolumn{1}{c}{\textbf{61.48}} & \multicolumn{1}{c|}{\textbf{100.11}} & \textbf{47.71} \\ \hline
w/o LN & \multicolumn{1}{c}{5.08} & \multicolumn{1}{c}{29.99} & \multicolumn{1}{c}{63.27} & \multicolumn{1}{c|}{103.58} & 49.16 \\ 
w/o truncation & \multicolumn{1}{c}{4.99} & \multicolumn{1}{c}{29.68} & \multicolumn{1}{c}{61.83} & \multicolumn{1}{c|}{100.13} & 47.99 \\
w/o sumNorm. & \multicolumn{1}{c}{5.37} & \multicolumn{1}{c}{30.20} & \multicolumn{1}{c}{63.14} & \multicolumn{1}{c|}{102.53} & 49.11 \\ \hline
w/o context & \multicolumn{1}{c}{5.17} & \multicolumn{1}{c}{30.15} & \multicolumn{1}{c}{62.54} & \multicolumn{1}{c|}{101.53} & 48.63 \\ \bottomrule[1pt]
\end{tabular}
\end{center}
\vspace{-2em}
\end{table}

\subsection{Ablation studies}
\textbf{Importance of Each Architecture Component}
{Table~\ref{Ablation-arch} summarizes the impact of key components in our model. Removing LSI or GSI increases the average MPJPE by 0.40 mm and 0.97 mm, respectively, confirming the complementary roles of fine-grained and holistic spatial modeling in our LGSI module. Excluding the entire LGSI module leads to a 0.44 mm performance drop, further validating its overall contribution. The AWA component also proves vital, with its removal causing a 1.57 mm increase. Similarly, eliminating the Traj or Pose module from the Traj-Pose structure increases the error by 0.51 mm and 4.20 mm, respectively, highlighting the effectiveness of our coarse-to-fine strategy. These results collectively demonstrate that optimal accuracy relies on the synergy of all architectural components.

\textbf{Ablation on the Architecture of the AWA module} 
Experimental results comparing various implementations of the AWA module are presented in Table~\ref{Ablation-AWI}. We first examine the effect of removing context-processing LN, which leads to a significant performance degradation in long-term prediction, with the MPJPE@1000 ms increasing from 100.11 mm to 103.58 mm. This result clearly demonstrates the importance of proper context normalization.
We then investigate the consequences of relaxing weight constraints by testing two alternative methods: one that removes the truncation of weight values to [0,1] (referred to as w/o truncation) and another that removes the normalization of weights by their sums (referred to as w/o sumNorm). Both approaches result in elevated MPJPE values, confirming the necessity of maintaining appropriate weight constraints.
Furthermore, we evaluate a context-free variant that uses learnable parameters as fixed weights (referred to as w/o context). This variant increases the average MPJPE by 0.92 mm compared with our full model, thereby validating the effectiveness of our context-based adaptive weighting mechanism in achieving optimal feature fusion performance.

\begin{table}[t]
\caption{Ablation on the Traj-Pose Module. Our Learned Pseudo-trajectory Outperforms Handcrafted Baselines Across all Time Steps on the HARPER Dataset.}
\vspace{-1.5em}
\label{Trajectorybaseline}
\begin{center}
\begin{tabular}{llllll}
\toprule[1pt]
\multicolumn{1}{c|}{\multirow{2}{*}{Method}} & \multicolumn{5}{c}{MPJPE   (mm)}                                                                                                      \\ \cline{2-6} 
\multicolumn{1}{c|}{}                        & \multicolumn{1}{c|}{100}  & \multicolumn{1}{c|}{400}   & \multicolumn{1}{c|}{700}   & \multicolumn{1}{c|}{1000}   & \multicolumn{1}{c}{Avg.}  \\ \hline
\multicolumn{1}{c|}{hip-based}               & \multicolumn{1}{c|}{4.95} & \multicolumn{1}{c|}{29.50} & \multicolumn{1}{c|}{61.96} & \multicolumn{1}{c|}{101.01} & \multicolumn{1}{c}{48.11} \\ \hline
\multicolumn{1}{c|}{mean-joint-based}        & \multicolumn{1}{c|}{5.02} & \multicolumn{1}{c|}{29.44} & \multicolumn{1}{c|}{61.55} & \multicolumn{1}{c|}{100.35} & \multicolumn{1}{c}{47.85} \\ \hline

\multicolumn{1}{c|}{{\textbf{ours}}}                    & \multicolumn{1}{c|}{{\textbf{4.94}}} & \multicolumn{1}{c|}{{\textbf{29.25}}} & \multicolumn{1}{c|}{{\textbf{61.48}}} & \multicolumn{1}{c|}{{\textbf{100.11}}} & \multicolumn{1}{c}{{\textbf{47.71}}} \\ 
\bottomrule[1pt]
\end{tabular}
\end{center}
\vspace{-2em}
\end{table}

\begin{table}[t]
\caption{Ablation on the Loss Combination. }
\label{lossfunc}
\vspace{-1em}
\renewcommand{\arraystretch}{1.2}
\begin{tabular}{c|c|ccccc}
\toprule[1pt]
Loss & $\lambda$ & 100 & 400 & 700 & 1000 & Avg. \\ \hline
$\mathcal{L}_{human}$ & - & \multicolumn{1}{l}{5.05} & \multicolumn{1}{l}{29.71} & \multicolumn{1}{l}{62.43} & \multicolumn{1}{l}{101.45} & \multicolumn{1}{l}{48.42} \\ \hline

$\mathcal{L}_{human}$+ $\mathcal{L}_{robot}$ & - & 5.19 & 30.64 & 63.91 & 103.67 & 49.62 \\ \hline
\multirow{5}{*}{\begin{tabular}[c]{@{}l@{}}$\mathcal{L}_{human}$+ $\mathcal{L}_{robot}$\\       +$\lambda\mathcal{L}_{reg}$\end{tabular}} & 10 & 4.98 & 29.31 & 62.60 & 103.24 & 48.66 \\
 & 50 & 5.05 & 29.91 & 62.58 & 101.36 & 48.52 \\
 & 100 & \textbf{4.94} & \textbf{29.25} & \textbf{61.48} & \textbf{100.11} & \textbf{47.71}  \\
 & 150 & 5.10 & 29.80 & 62.44 & 101.27 & 48.45 \\
 & 200 & 5.06 & 29.36 & 61.90 & 101.20 & 48.10 \\ \bottomrule[1pt]
\end{tabular}
\vspace{-1em}
\end{table}

\textbf{Ablation on the Traj-Pose Module}
To evaluate the effectiveness of the pseudo-trajectory design in our Traj-Pose Module, we conduct ablation experiments by comparing our learned pseudo-trajectory against two handcrafted baselines: (1) a hip-based baseline that uses only the hip joint feature as a global motion proxy, and (2) a mean-joint-based baseline that uses the average of all joint features. 
As Table~\ref{Trajectorybaseline} shows, our method achieves the lowest average MPJPE, indicating that learning to aggregate joint information yields better global motion representation than handcrafted strategies.
}

\textbf{Ablation of the Loss Design}
The contribution of each component in our loss function is shown in Table ~\ref{lossfunc}. Because our dual-branch network is designed to simultaneously predict human motion and robot motion, $\mathcal{L}_{human}+\mathcal{L}_{robot}$ is a natural choice for supervising both tasks. However, this combination yields suboptimal results, even when only $\mathcal{L}_{human}$ is used. To address this issue, we introduce $\mathcal{L}_{reg}$, which supervises the coefficients generated by the AWA module, enabling better fusion of local and global features. The hyperparameter $\lambda$ controls the weight of $\mathcal{L}_{reg}$, balancing its influence on the overall loss. 
The best performance is achieved by combining $\mathcal{L}_{human}+\mathcal{L}_{robot}+\lambda\mathcal{L}_{reg}$ with an appropriately chosen value of $\lambda$. This highlights the importance of jointly optimizing task-specific losses and feature fusion regularization. 

\section{Conclusion}
In this study, we utilize the interaction information between the human and the robot to improve the performance of human motion prediction during human-robot collaboration. The following conclusions can be drawn: 
First, the proposed robot-aware human motion prediction model effectively captures the dynamic interaction between the human and robot, enabling more accurate and contextually reasonable predictions during collaborative tasks. 
Second, incorporating interaction information from both local and global spatial dimensions significantly improves the model's contextual understanding of human-robot collaboration scenarios. 
Third, the context-based adaptive weighting mechanism flexibly aggregates motion representations, further boosting prediction performance.

Although our method achieves state-of-the-art results, it still shows higher errors on certain collision-related actions involving abrupt transitions. This highlights a potential limitation in handling irregular motion patterns. 
Future work may explore transition-aware modeling and targeted data augmentation to enhance robustness in such cases. 
In addition, while our current framework focuses on single human--robot pairs, it is inherently extensible to multi-agent scenarios. The dual-branch design and interaction modeling module can be generalized to an $N$-branch architecture, where each branch predicts one agent's motion while incorporating the other $N{-}1$ agents as contextual inputs. Such a design enables scalable and structured interaction modeling, laying the groundwork for future extensions to more complex collaborative settings.

\bibliographystyle{IEEEtran}
\bibliography{IEEEabrv,bibliography_citedrive}

\vspace{-19pt}
\begin{IEEEbiography}[{\includegraphics[width=1in,height=1.25in]{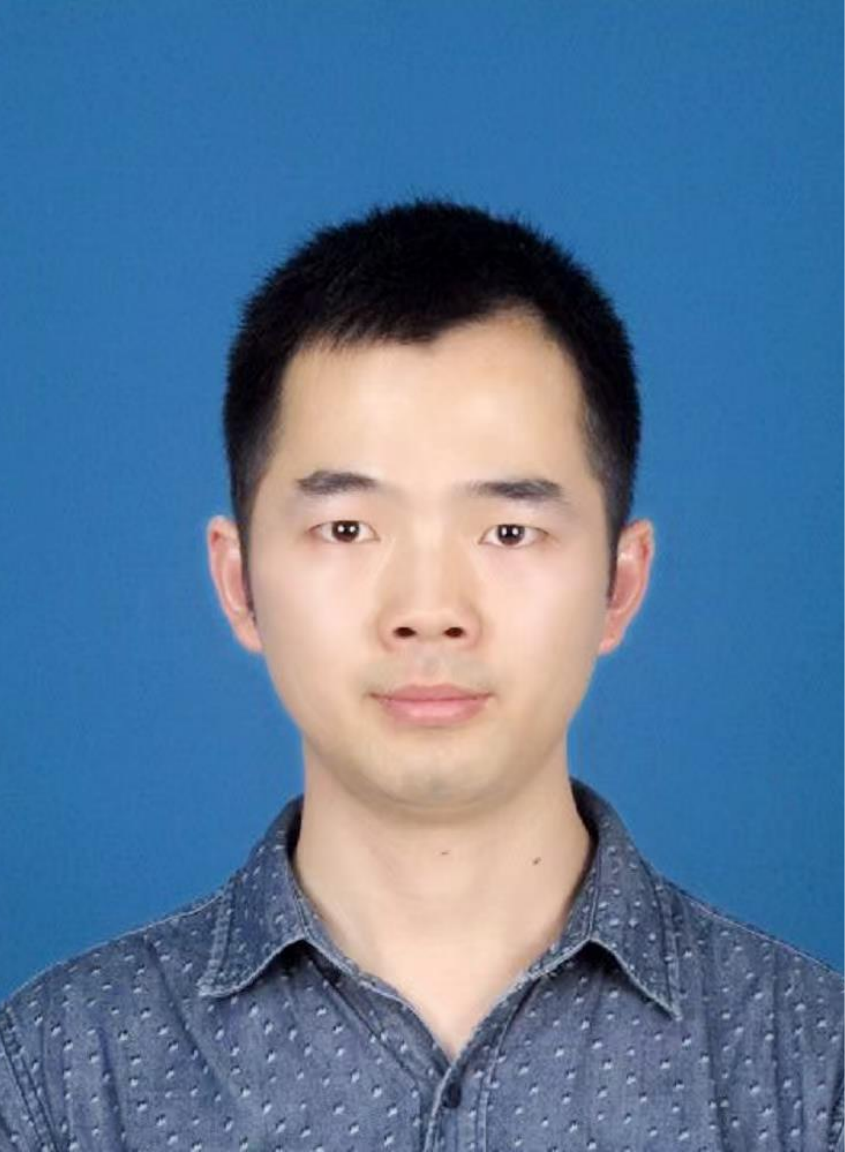}}]{Mengyuan Liu}
Mengyuan Liu received his Ph.D. under the supervision of Prof. Hong Liu at the School of Electrical Engineering and Computer Science (EE\&CS), Peking University (PKU), China. He subsequently worked as a research fellow with Prof. Junsong Yuan and Prof. Kai-Kuang Ma at the School of Electrical and Electronic Engineering, Nanyang Technological University (NTU), Singapore. Currently, he is an Assistant Professor at Peking University, Shenzhen Graduate School. His research primarily focuses on human-centric perception and human-robot interaction. His work has been published in leading conferences and journals, including NeurIPS, CVPR, ICRA, T-IP, T-CSVT, T-MM, and PR. In addition, he actively serves as a reviewer for several top-tier international conferences and journals, such as NeurIPS, CVPR, T-IP, T-RO, and T-PAMI.
\end{IEEEbiography}
\vspace{-15pt}

\begin{IEEEbiography}[{\includegraphics[width=1in,height=1.25in]{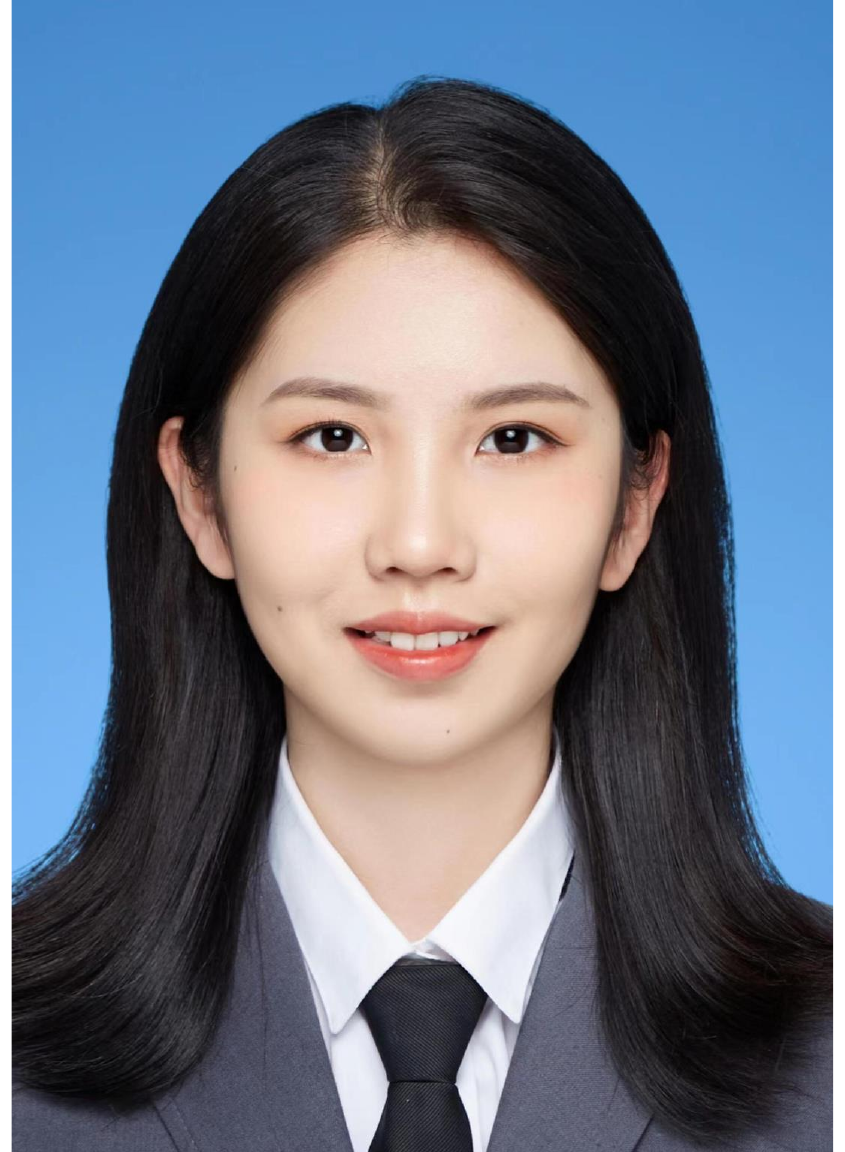}}]{Yangting Lin}
Yangting Lin received the B.S. degree from the Beijing University of Post and Telecommunications. She is currently pursuing the M.S. degree in the State Key Laboratory of General Artificial Intelligence, Peking University, Shenzhen Graduate School. Her research interests include human motion analysis, computer vision and machine learning. 
\end{IEEEbiography}

\vspace{-15pt}
\begin{IEEEbiography}[{\includegraphics[width=1in,height=1.25in]{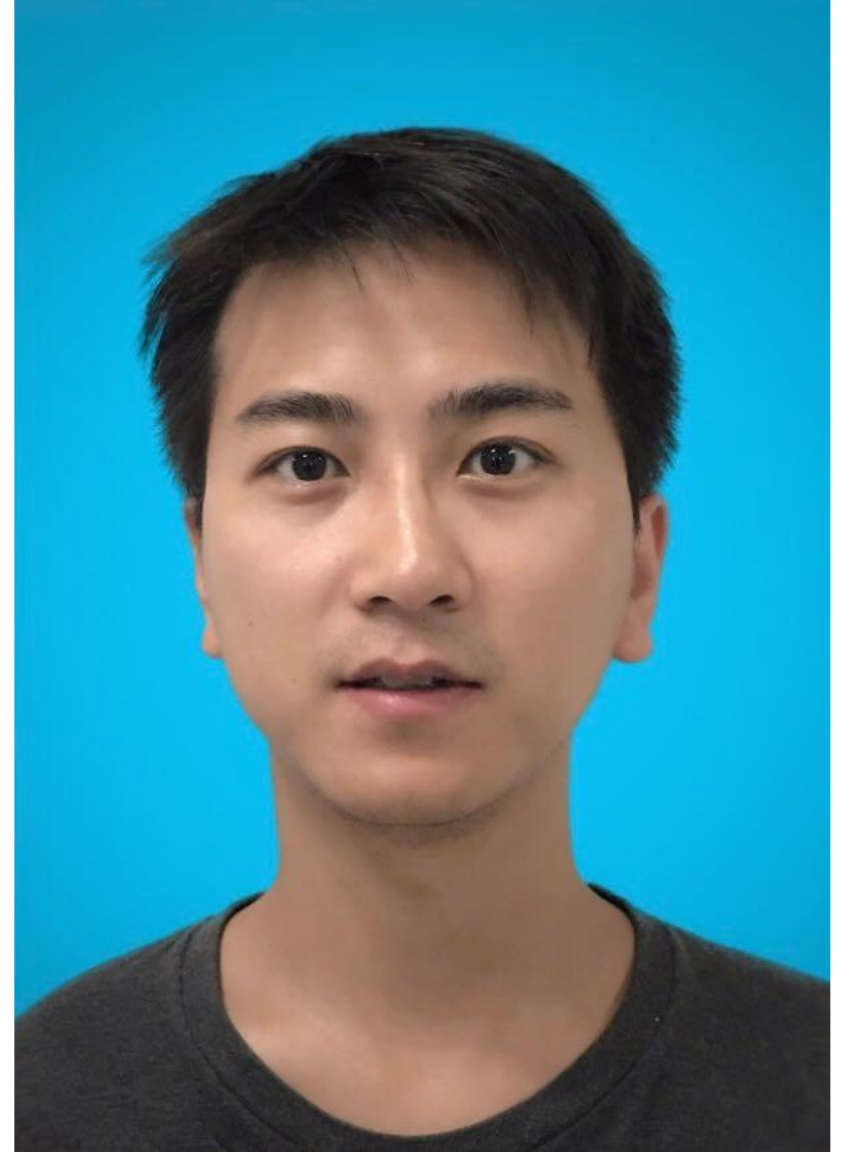}}]{Qiongjie Cui}
Qiongjie Cui received the B.S. degree from Northwestern Polytechnical University, Xi'an, China, in 2015, and the Ph.D. degree from Nanjing University of Science and Technology, Nanjing, China, in 2021. He is currently an Associate Professor with the National Key Laboratory of Human-Machine Hybrid Augmented Intelligence, Xi'an Jiaotong University, Xi'an, China. Previously, he was a Research Fellow with Singapore University of Technology and Design, Singapore. His research interests include 3-D computer vision and human motion analysis, prediction, reconstruction, and capture. He has published more than 30 papers in these areas. He serves on the Editorial Board of \emph{Cyborg and Bionic Systems} and has served as an Area Chair for PRCV, ICIG, and ICME.
\end{IEEEbiography}

\vfill

\end{document}